\documentclass{article}
\usepackage[T1]{fontenc}
\usepackage{iclr2027_conference,times}
\usepackage{amsmath,amssymb,amsfonts,mathtools,bm,amsthm}
\usepackage{booktabs,tabularx,array}
\usepackage{microtype,xcolor,colortbl,tikz,enumitem,graphicx}
\usepackage{placeins,needspace,flafter,float}
\usepackage{capt-of}
\usepackage[bottom]{footmisc}
\usepackage[colorlinks=true,allcolors=blue!55!black]{hyperref}
\usepackage[nameinlink,capitalize]{cleveref}

\newcommand{\vd}{\bm{\delta}}
\newcommand{\vk}{\bm{k}}
\newcommand{\vv}{\bm{v}}
\newcommand{\vq}{\bm{q}}
\newcommand{\vS}{\bm{S}}
\newcommand{\vI}{\bm{I}}
\newcommand{\Diag}{\operatorname{Diag}}
\newcommand{\R}{\mathbb{R}}
\newcommand{\looptopology}[4]{\ensuremath{[#1\textnormal{L}\to#2\textnormal{L}\,(\times#3)\to#4\textnormal{L}]}}

\definecolor{paperTeal}{HTML}{007F78}
\definecolor{paperGreen}{HTML}{2E7D32}
\definecolor{paperNavy}{HTML}{243746}
\renewcommand{\arraystretch}{1.12}


\iclrfinalcopy
\usepackage{etoolbox}
\makeatletter
\patchcmd{\@maketitle}{\lhead{Published as a conference paper at ICLR 2027}}{\lhead{Preprint}}{}{%
  \PackageError{main}{Could not patch the running header of the title block}{}}
\makeatother
\usepackage{marvosym}    % \Letter (envelope) for the corresponding-author mark
\usepackage{fontawesome5} % \faGithub / \faGlobe beside the code and project-page links
\title{Loopback Transformer: Towards a Unified View of Residual Looped Transformers}
\author{Xuan Ouyang\textsuperscript{1,*}, Zefan Cai\textsuperscript{1,*}, Chenlong Zhang\textsuperscript{2}, Cheng Luo\textsuperscript{3}, Ke Wan\textsuperscript{4}, Junjie Hu\textsuperscript{1,\Letter} \\[3pt]
  {\normalfont \textsuperscript{1}University of Wisconsin--Madison \quad \textsuperscript{2}University of California, Santa Barbara} \\[1pt]
  {\normalfont \textsuperscript{3}California Institute of Technology \quad \textsuperscript{4}Microsoft} \\[3pt]
  {\normalfont\small \textsuperscript{*}Equal contribution \quad \textsuperscript{\Letter}\,Corresponding author}}
\hypersetup{pdftitle={Loopback Transformer: Towards a Unified View of Residual Looped Transformers},
  pdfauthor={Xuan Ouyang, Zefan Cai, Chenlong Zhang, Cheng Luo, Ke Wan, Junjie Hu}}
\begin{document}
\raggedbottom
\maketitle
\begingroup
\renewcommand{\thefootnote}{\Letter}
\footnotetext[0]{Correspondence to Xuan Ouyang, Zefan Cai, and Junjie Hu
(\href{mailto:xouyang7@wisc.edu}{xouyang7@wisc.edu},
\href{mailto:zefncai@gmail.com}{zefncai@gmail.com},
\href{mailto:jhu@cs.wisc.edu}{jhu@cs.wisc.edu}).}
\endgroup

\begin{abstract}
Looped Transformers reduce parameter count by repeatedly applying shared blocks, but parameter sharing alone does not determine how the results of repeated computation are reused. We take the depth-memory view of residual connections and analyze which representation is stored after each block application and how later reads weight it. A full block output contains both its input and its new update, so weighting that output also reweights the history already carried by the input. Reading individual updates instead, with the shared stack's input kept on a separate path, gives separate controls over input strength, total update weight, and allocation across updates. The Loopback Transformer implements these controls: geometric decay sets the total update weight and supplies a recency prior, and attention learns how to allocate that weight. On ClimbMix, Loopback reaches lower validation loss than standard Transformer baselines with 40--75\% of their Transformer parameters (excluding token embeddings and the output head) at reference scales from 191M to 3B, with training tokens and executed depth matched. It also has higher mean accuracy over thirteen downstream tasks at 1.4B and 3B. At 191M, 374M, and 778M, it has the lowest validation loss and the highest thirteen-task mean among the standard Transformer and seven looped residual designs.
\end{abstract}

\begin{center}
\vspace{-6pt}
\small
\faGithub\ \textbf{Code:} \href{https://github.com/yancyou/Loopback}{github.com/yancyou/Loopback}
\qquad
\faGlobe\ \textbf{Project page:} \href{https://yancyou.github.io/Loopback_page/}{yancyou.github.io/Loopback\_page}
\end{center}

% Figure 1, right after the abstract (2026-10-27): generated by figure/intro_fig1/fig_intro_final.py (outside this
% repository) from the values of tab:downstream-main (191M, and 374M / 778M at 32 / 48 layers) and tab:scale-large
% (1.4B, 3B). Three panels: (a) validation loss and (c) the thirteen-task mean against the standard Transformer's
% training FLOPs (three times the forward count of app:compute, times tokens; Loopback's reads, +0.02-0.17% at these
% shapes, are not drawn -- tab:compute lists the extra FLOPs), and (b) the same losses against Transformer parameters.
% The script checks that no label touches another label, a curve, a marker or an axis; regenerate fig_intro_compute.pdf
% after any change to those tables. [H] keeps the figure under the abstract on page 1, and the caption is sized to the
% space left there: if the abstract grows, shorten the caption (or the figure) rather than let the figure slip to page 2.
\begin{figure}[H]
\centering
\vspace{-2pt}
\includegraphics[width=\linewidth]{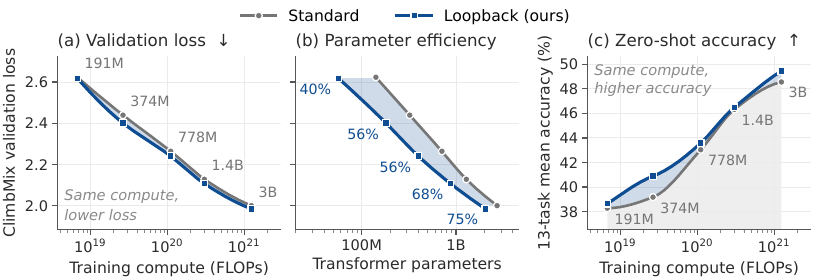}
\vspace{-6pt}
\caption{\textbf{Loopback versus the standard Transformer from 191M to 3B.} Each pair uses the same token budget
and executed depth at the 20-TPP setting (tokens per reference-model parameter). (a)~Validation loss and
(c)~thirteen-task mean accuracy against the standard Transformer's estimated training FLOPs; (b)~the same losses against
Transformer parameters, where percentages give Loopback's share of the standard Transformer's. Lines interpolate the
points.}
\label{fig:intro-compute}
\end{figure}

\section{Introduction}
\label{sec:introduction}

Looped Transformers reduce parameter count by repeatedly applying shared blocks
to an evolving hidden representation \citep{dehghani2018universal,lan2019albert}.
Sharing separates computational depth from the number of distinct blocks and
supports latent reasoning and inference-time scaling
\citep{saunshi2025reasoning,geiping2026scaling}. Each visit applies the same transformation to a different hidden state and so produces a different update. Parameter sharing fixes which transformation is reused; it leaves open how the updates from successive visits are stored, carried forward, and used.

Residual connections turn depth into a form of memory: each layer writes new
computation into a residual stream carried forward to later layers. Standard
residual addition accumulates block updates with fixed unit
coefficients~\citep{he2016deep}, whereas Attention Residuals (AttnRes) learns how to
weight these updates and the input~\citep{team2026attention}.
Hyper-Connections (HC) and Hyperloop introduce learned pathways across residual
streams or loop visits~\citep{zhu2025hyper,zeitoun2026hyperloop}.

This depth-memory view~\citep{team2026attention} connects residual design to
linear attention and related recurrent models. Linear attention expresses
attention over token history as writes to and reads from a recurrent
memory~\citep{katharopoulos2020transformers}; RetNet and Mamba-2 additionally
control how that memory is retained across
steps~\citep{sun2023retentive,dao2024transformers}. We use this
\emph{write--transport--read} perspective to analyze repeated computation
along depth. Here, each block produces a new write from a representation
assembled from earlier writes, so the memory rule also influences subsequent
computation. We therefore examine what is stored and how memory transitions
and readout determine each update's contribution.

Seen this way, common residual designs tie together quantities that could be
controlled separately. First, a full block output contains both its input and
its new update. Weighting that output therefore also weights the history
carried by the input. Storing updates separately lets the reader assign a
coefficient directly to each update. Second, a joint softmax over the original
input and stored updates makes them compete for a fixed total weight. Ordinary
addition instead fixes the input coefficient but assigns every update weight
one, so total update weight grows with the number of block applications. We
therefore want separate controls over \emph{input strength}, \emph{total update
weight}, and \emph{allocation across updates}.

The \textbf{Loopback Transformer} is built around these three controls. It
stores individual block updates rather than full block outputs and keeps the shared
stack's input on a separate path with coefficient one. Geometric decay assigns
a baseline weight to each update. A softmax reader redistributes their total
weight according to update content, using the decay weights as a recency prior.
This separates total update weight from allocation without changing the
shared blocks or their token-axis attention.

We compare against the standard Transformer, which uses distinct parameters at every layer (labeled \emph{Standard} in tables and figures), and against seven looped residual designs, always at the same budget of training tokens and the same executed depth.
On ClimbMix \citep{diao2026nemotron}, Loopback reaches lower validation loss and a higher thirteen-task mean than Standard at 191M, 374M, 778M, 1.4B, and 3B while storing 40--75\% of its Transformer parameters (\cref{tab:downstream-main,tab:ablation,tab:scale-large,fig:intro-compute}). At 191M, 374M, and 778M, it also has the lowest validation loss and the highest thirteen-task mean of all nine designs (\cref{tab:downstream-main}).

\Needspace{5\baselineskip}   % keep the lead-in with its list (Figure 1 under the abstract moved the introduction down)
We make three contributions:
\begin{itemize}[leftmargin=1.5em,itemsep=1pt,topsep=3pt,parsep=0pt]
    \item We connect residual designs with linear attention and related recurrent models by adapting their memory-update rules from token sequences to network depth. The resulting framework describes what each block writes and how its contribution is transported and read.
    \item Within this framework, we design the Loopback Transformer, which adds a weighted sum of block updates to the original loop input. Geometric decay keeps the total update weight bounded as visits increase at fixed stack size. Attention adjusts each update's weight according to its content while preserving the total.
    \item On ClimbMix, Loopback reaches lower validation loss and higher zero-shot task means than matched-depth standard Transformers at five scales from 191M to 3B, with 40--75\% of their Transformer parameters.
\end{itemize}

\section{Preliminaries}
\label{sec:preliminaries}
\label{sec:messages}

\subsection{Looped Transformer}

A looped Transformer shares parameters across depth by repeatedly applying
the same sequence of Transformer blocks. Each block contains causal token
attention followed by an MLP, with a residual connection around each
computation. We call this shared sequence a \emph{stack}, consisting of $K$
distinct blocks with transformations $f_0,\ldots,f_{K-1}$.

We adopt Huginn's prefix--middle--coda layout
\citep{geiping2026scaling}, with $P$ prefix blocks, a shared middle
stack of $K$ blocks repeated for $R$ loop iterations, and $C$ coda
blocks. Following \citet{zeitoun2026hyperloop}, we denote this topology
by $\looptopology{P}{K}{R}{C}$, where the suffix \textnormal{L}
denotes layers, counting each block application as one layer of executed
depth. The model stores $P+K+C$ distinct blocks
and executes $P+RK+C$ block applications. Let $\bm h_0$ be the prefix
output entering the shared stack. Within this stack,
$r\in\{0,\ldots,R-1\}$ indexes loop iterations, and
$k\in\{0,\ldots,K-1\}$ identifies a block's position within each
iteration. The index $s=rK+k\in\{0,\ldots,RK-1\}$ counts block
applications across all iterations. At application $s$, we apply the
shared block $f_k$, where $k=s\bmod K$.

To describe how residual connections carry computation across depth,
we distinguish a block's full output from the update it produces.
Let $\bm h_s$ denote the aggregated hidden representation before
application $s$, and let $\bm x_s$ be the input passed to block $f_k$.
A single-stream model passes one hidden representation between blocks,
so $\bm x_s=\bm h_s$; a multi-stream model may form $\bm x_s$ by mixing
several parallel hidden states.
Aggregating the block's result with stored computation produces
the next readout $\bm h_{s+1}$.
The \emph{full block output} is $f_k(\bm x_s)$, and the
\emph{block update} is
\begin{equation}
    \vd_s=f_k(\bm x_s)-\bm x_s.
    \label{eq:update}
\end{equation}
We suppress batch and token indices; each token's block update is
in $\R^d$, where $d$ is the hidden width.
The block update measures the change produced by the block itself,
before residual aggregation.

\subsection{Residual aggregation}

We review ordinary residual addition, Attention Residuals, and
Hyper-Connections to examine how block updates contribute to later
computation. These designs accumulate updates directly, weight stored
computation through attention, or maintain multiple residual streams.

\textbf{Ordinary residual addition}~\citep{he2016deep} takes $\bm x_s=\bm h_s$ and uses a fixed coefficient of one to accumulate each update:
\begin{equation}
    \bm h_{s+1}=\bm h_s+\vd_s
              =\bm h_0+\sum_{i=0}^{s}\vd_i.
    \label{eq:plain-sum}
\end{equation}
% Each update retains coefficient one. 

\textbf{Attention Residuals (AttnRes)}~\citep{team2026attention} learns
content-dependent weights over the input and stored computation.
Before block application $s$, its block-level formulation reads the
sources $\bm y_0=\bm h_0$ and $\bm y_{i+1}=\vd_i$ for $0\le i<s$:
\begin{equation}
    \bm h_s=\sum_{i=0}^{s}\pi_{s,i}\bm y_i,\qquad
    \pi_{s,i}=\operatorname{softmax}_i
       \bigl(\vq_s^\top\operatorname{RMSNorm}(\bm y_i)\bigr).
    \label{eq:attnres}
\end{equation}
The learned query $\vq_s\in\R^d$ scores normalized sources, while the read
combines their original values.

\textbf{Hyper-Connections (HC)}~\citep{zhu2025hyper} carries
$n$ parallel $d$-dimensional hidden states per token through the
network, forming $n$ residual streams. Before block application $s$,
these states are stored side by side in
$\bm H_s\in\R^{d\times n}$.
At block level, HC combines these states into one input and
distributes the block update back to the streams:
\begin{equation}
    \bm x_s=\bm H_s\bm p_s,\qquad
    \bm H_{s+1}=\bm H_s\bm W_s+
       \vd_s\bm b_s^\top.
    \label{eq:hc}
\end{equation}
Here $\bm p_s\in\R^n$ combines the states into the block input,
$\bm W_s\in\R^{n\times n}$ mixes the existing states, and
$\bm b_s\in\R^n$ controls how much of the block update each
stream receives. These weights are learned and may depend on
$\bm H_s$.

\subsection{Linear attention and recurrent memory}

Linear attention \citep{katharopoulos2020transformers} summarizes
token history in a recurrent memory. Each new token adds information
to this memory, from which a query reads an output.
A scalar-decay version of the unnormalized recurrence is
\begin{equation}
    \vS_t=\gamma_t\vS_{t-1}+\vv_t\vk_t^\top,
    \qquad \bm o_t=\vS_t\vq_t.
    \label{eq:background-memory}
\end{equation}
Here $t$ indexes tokens, $\vv_t\in\R^d$ is the value, and
$\vk_t,\vq_t\in\R^m$ are the key and query, with feature dimension $m$.
The matrix $\vS_t\in\R^{d\times m}$ stores the accumulated writes.
The factor $\gamma_t$ scales the previous memory,
$\vv_t\vk_t^\top$ writes the current value, and
$\vq_t$ reads the unnormalized output $\bm o_t$.
With feature maps
absorbed into the keys and queries, additive linear attention uses
$\gamma_t=1$ and divides the read by the positive normalizer
$\bm z_t^\top\vq_t$, where $\bm z_t=\sum_{i\le t}\vk_i$.
Related recurrent models modify transport: RetNet uses fixed decay
\citep{sun2023retentive}, while Gated Linear Attention (GLA) and Mamba-2 make it
input dependent \citep{yang2023gated,dao2024transformers}.
Along depth, each token position can maintain a memory of earlier
block computations.
% \junjie{this sentence is not very clear to me}
After a block runs, its result for that token updates the memory,
whose readout helps form the next block's input.
Attention inside each block still operates over tokens.
% \junjie{the last few sentences are too abstract and not very clear to me. }

\section{Residual connections as depth memory}
\label{sec:method}

Following the depth-memory view \citep{team2026attention}, we describe
residual connections and recurrent memory by the same write,
transport, and read operations. We first compare full block outputs
with individual block updates as the stored value, then expand the
recurrence to see how the original input and earlier block updates
enter later hidden states. The expansion is what we use to compare
how existing residual designs reuse earlier computation, and it leads
to the design of Loopback.

\subsection{A common memory interface}
\label{sec:unified-interface}

To trace how block results contribute to later computation, we describe
residual aggregation through a common memory formulation.
For each token, let $\bm M_s\in\R^{d\times m_s}$ hold the representations
retained before block application $s$, where $m_s$ counts the
$d$-dimensional vectors used to represent memory.
The block input is $\bm x_s=\bm M_s\bm p_s$, with
$\bm p_s\in\R^{m_s}$. Its result supplies a value $\vv_s\in\R^d$
to write into memory, such as the full block output or the block update:
\begin{equation}
 \begin{aligned}
    \bm M_{s+1}&=\bm M_s\bm W_s+\vv_s\bm b_s^\top,\\
    \bm h_{s+1}&=\bm M_{s+1}\bm r_s.
 \end{aligned}
    \label{eq:two-axis-memory}
\end{equation}
The transport matrix $\bm W_s\in\R^{m_s\times m_{s+1}}$ scales or
mixes the existing representations. The write term
$\vv_s\bm b_s^\top$ adds the current value to memory, with
$\bm b_s\in\R^{m_{s+1}}$ specifying its weight in each memory
component. The read combines the updated representations using
$\bm r_s\in\R^{m_{s+1}}$ to produce $\bm h_{s+1}\in\R^d$.
Thus, $\bm p_s$ forms the input before the block runs, while
$\bm r_s$ reads the memory after it is updated; these weights may
depend on content. Single-stream models pass $\bm h_{s+1}$ to the
next block. Multi-stream models form the next input from
$\bm M_{s+1}$ using $\bm p_{s+1}$.
Unrolling this recurrence follows each written value through
subsequent transport and read operations and gives its weight
in a later readout, so different residual designs can be compared
even when their memory representations differ.
\Cref{tab:residual-families} summarizes their choices.

The memory can store past values separately or combine them in a
fixed-width state. An explicit history bank grows as new values
are added, while recurrent memory updates a state of fixed size.
Along depth, each block application supplies a value $\vv_i$.
Let $\vS_i$ denote the accumulated writes after application $i$,
starting from $\vS_{-1}=0$, and let
$\vq_s$ be the query used to read after application $s$.
With feature maps absorbed into the write vectors and queries,
the unnormalized recurrence and its expansion are
\begin{subequations}
\label{eq:depth-memory-kernel}
\begin{align}
    \vS_i&=\vS_{i-1}\bm W_i+\vv_i\bm b_i^\top,\qquad\vS_{-1}=0,
       \label{eq:memory}\\
    \bm o_s&=\vS_s\vq_s=\sum_{i=0}^{s}
       (\bm b_i^\top\bm W_{i+1}\cdots\bm W_s\vq_s)\vv_i.
       \label{eq:memory-unroll}
\end{align}
\end{subequations}
The second line follows by repeatedly substituting the first: each value is
written by $\bm b_i$, transported through subsequent $\bm W_j$, and read by
$\vq_s$. Empty products are identity matrices. In the common interface,
$\bm M_i$ denotes the memory before application $i$.
For the zero-initialized case $\bm M_0=0$, the correspondence is
$\bm M_i=\vS_{i-1}$, $\bm M_{i+1}=\vS_i$, and $\bm r_i=\vq_i$.
This initializes the value memory, not the network input; an input
path can be kept separately or included in a nonzero $\bm M_0$.
For a general initial state, unrolling the common recurrence also
contributes $\bm M_0\bm W_0\cdots\bm W_s\bm r_s$ to the readout.
The general block-update expansion below retains this initial-state
contribution.
When $\vv_i=\vd_i$, these coefficients directly weight individual
block updates. A full block output also contains its input, so
its coefficient affects earlier computation as well.

% Author-supplied vector artwork; panel (c) combines memory update and output read.
% Introduce the three pathways after Section 3.1, before the value-choice analysis.
\begin{figure}[!htbp]
\centering
\includegraphics[width=\linewidth]{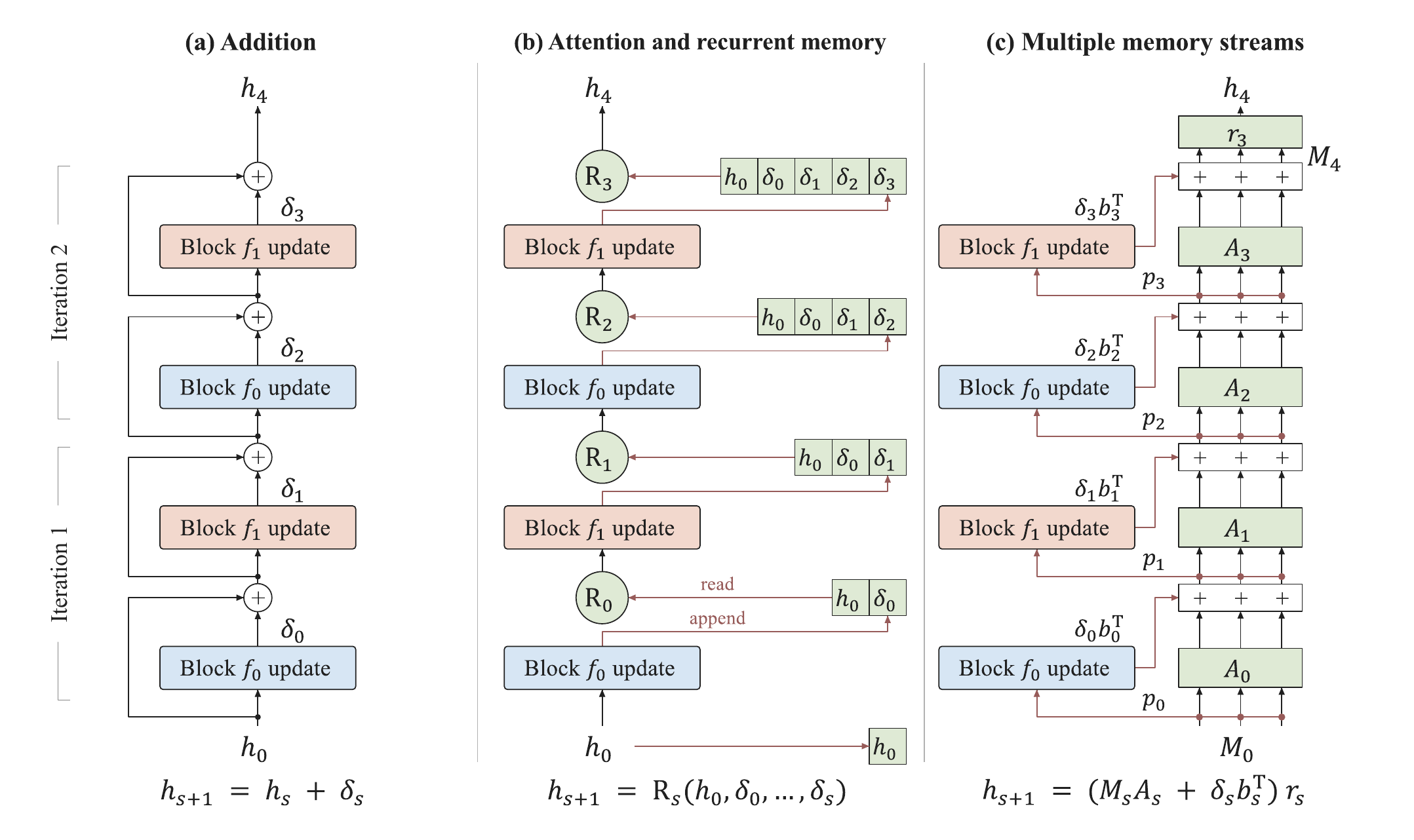}
\caption{\textbf{Same shared blocks, three residual pathways.}
Repeated colors denote shared blocks emitting block updates.
(a) Addition. (b) $\mathcal R_s$ aggregates the input and updates computed so far
using explicit history or a fixed-size recurrent memory.
(c) Multiple streams with input read, transport, write, and final readout.}
\label{fig:unified-residual}
\end{figure}
\FloatBarrier

\subsection{Outputs versus updates}
\label{sec:block-history}
\label{sec:delta}
\label{sec:loop-motivation}

A full block output $f_k(\bm x_s)=\bm x_s+\vd_s$ contains both
the current block update and earlier computation carried by its input.
A read weight on this output therefore applies to both.
Storing $\vd_s$ separately lets the read assign a coefficient
directly to that block update.

Consider two successive full block outputs under ordinary residual
addition: $\bm h_1=\bm h_0+\vd_0$ and
$\bm h_2=\bm h_0+\vd_0+\vd_1$.
Let $\rho\in[0,1]$ be the read weight on $\bm h_1$, with
the remaining weight $1-\rho$ assigned to $\bm h_2$.
Their weighted read is
\[
    \rho\bm h_1+(1-\rho)\bm h_2
       =\bm h_0+\vd_0+(1-\rho)\vd_1.
\]
Because $\vd_0$ appears in both stored outputs, its coefficient
is always one. The update $\vd_1$ appears only in $\bm h_2$,
so its coefficient is $1-\rho$.
More generally, a nonnegative normalized read over
ordinary additive states assigns nonincreasing coefficients to successive
updates (\cref{app:cumulative-states}). This ordering need not hold for
reweighted trajectories. Reading updates separately removes the
cumulative-sum restriction.

The example shows how a stored output carries updates already
present in earlier states. In the common memory formulation,
this occurs through the input term in
$f_k(\bm x_s)=\bm x_s+\vd_s$.
To make its effect explicit, let the binary indicator $\chi$
specify the stored value: $\chi=0$ selects the full block output,
and $\chi=1$ selects the block update.
The definition in \cref{eq:update} then gives
\begin{equation}
    \vv_s^{(\chi)}=f_k(\bm x_s)-\chi\bm x_s
       =\vd_s+(1-\chi)\bm x_s,\qquad\chi\in\{0,1\}.
    \label{eq:delta-choice}
\end{equation}
Using $\bm x_s=\bm M_s\bm p_s$ in \cref{eq:two-axis-memory} gives
\begin{equation}
    \bm M_{s+1}=\bm M_s\underbrace{
       [\bm W_s+(1-\chi)\bm p_s\bm b_s^\top]}_{\bm A_s}
       +\vd_s\bm b_s^\top.
    \label{eq:delta-transport}
\end{equation}
Here $\bm A_s$ is the effective transport when the recurrence
is expressed in terms of block updates. With full-output writes,
the additional term carries the input back into memory, since
$\bm M_s\bm p_s\bm b_s^\top=\bm x_s\bm b_s^\top$.
With block-update writes, this term vanishes and
$\bm A_s=\bm W_s$. Constraints on $\bm W_s$ alone therefore
need not hold for the effective transport $\bm A_s$.
The block update $\vd_s$ still depends on the input.
An unrestricted transport operator can absorb the extra term;
the distinction concerns parameterization and control, not a general
expressivity advantage.

\Needspace{10\baselineskip}
\subsection{Input and update weights}
\label{sec:families}
\label{sec:composition}

Expanding \cref{eq:delta-transport} separates the original input from all
subsequent block updates, including those contained in full block outputs.
For $\bm M_0=\bm h_0\bm a_0^\top$, the vector
$\bm a_0\in\R^{m_0}$ specifies how the input is placed in memory
(all ones, for example, copy it into every stream). The read becomes
\begin{align}
    \bm h_{s+1}&=a_{s+1}\bm h_0+\sum_{i=0}^{s}c_{s,i}\vd_i,
       \label{eq:message-expansion}\\
    c_{s,i}&=\bm b_i^\top\bm A_{i+1}\cdots\bm A_s\bm r_s,\qquad
    a_{s+1}=\bm a_0^\top\bm A_0\cdots\bm A_s\bm r_s.
       \label{eq:history-coefficients}
\end{align}
Each $c_{s,i}$ combines write, subsequent transport, and final read;
$a_{s+1}$ includes the input carried inside stored values. Empty products are identity matrices.
All quantities are evaluated along the actual nonlinear trajectory: this is
a value decomposition, not a linear model or Jacobian. It applies headwise;
value projections require operator-valued coefficients
(\cref{app:framework-details}).

For nonnegative coefficients, we can separate their total from their
distribution. Suppressing the destination index and taking
$B=\sum_i c_i>0$, write
\begin{equation}
    \bm h=a\bm h_0+B\sum_i\pi_i\vd_i,\qquad
    \pi_i=c_i/B,\quad\sum_i\pi_i=1.
    \label{eq:separated-controls}
\end{equation}
Here $a$ is the input coefficient, $B$ is the total update weight,
and $\pi_i$ specifies the fraction assigned to update $i$.
The total $B$ sums coefficients; it is not a representation norm.
Ordinary addition fixes $a=1$ and gives every update coefficient
one, so $B=s+1$ grows with the number of block applications.
A joint softmax over the input and block updates instead enforces
$a+B=1$: increasing the total weight on updates reduces the
input weight.
% \junjie{Loopback is only introduced in detail later. Readers may get confused about Loopback's details when reading this paragraph.}

\subsection{Residual and recurrent models}
\label{sec:depth-linear}

\Cref{tab:residual-families} contrasts explicit history, recurrent compression,
and stream transport. Addition starts from $\bm M_0=\bm h_0$.
AttnRes and Multi-Head Attention Residuals (MHAR)~\citep{luo2026multi} use
explicit history banks; the block-level formulations in the table append
block updates.
Our evaluated adapters use
Block AttnRes and a loop-state MHAR reader with an outer residual blend;
their sources and placements are specified in \cref{app:190m-evaluation}.
Manifold-constrained Hyper-Connections (mHC)~\citep{xie2025mhc} constrains
transport by applying Sinkhorn normalization to a learned matrix $\bm U_s$.
Hyperloop's transport is
diagonal, and iHC\footnote{\label{fn:ihc-source}Identity
Hyper-Connections; see the \href{https://huggingface.co/tencent/Hy4-preview}{official Hy4-preview model card}.}
uses identity transport.

For recurrent memory, $\gamma_s$ scales the entire previous state, while
$\Diag(\bm g_s)$ scales its components separately; $\Diag$ forms a
diagonal matrix from a vector. RetNet uses a fixed $\gamma$.
In normalized linear attention, $\bm z_{s+1}=\sum_{i=0}^{s}\vk_i$
accumulates keys and $\bm z_{s+1}^\top\vq_s>0$ is required.
Our evaluated linear-attention adapters omit this kernel normalizer and use
$\bm r_s=\vq_s$. GLA uses key-side gating, while Mamba-2 couples its scalar
decay and write scale through $\gamma_s=\exp(-\alpha\tau_s)$, $\alpha,\tau_s>0$.
The delta-rule families use $\eta_s$ to control a correction of the
memory's prediction of the new value; this correction is distinct from
the block update $\vd_s$. RWKV-7 uses a removal direction
$\bm\kappa_s$ with elementwise strength $\bm a_s$, where $\odot$
denotes elementwise multiplication.
Mesa additionally tracks inverse covariance; our LaCT-inspired reader
updates from the stored prefix. These extra states and the remaining
implementation conventions are detailed in \cref{app:operator-mapping}.

\begin{table}[!htbp]
\hypersetup{allcolors=black}
\setlength{\abovecaptionskip}{3pt}
\caption{Write, transport, and read operations for residual and attention memory,
with $\bm M_s\in\R^{d\times m_s}$,
applied per head where needed. Feature construction, output gates and normalization,
feedthrough paths, and outer residual blends are not shown. The rows describe
memory operations, not complete published architectures.}
\label{tab:residual-families}
\label{tab:families}
\centering
% W carries the supplied values; A in the delta expansion may also include
% an input-write term. All matrices below use value-by-column orientation.
\begingroup
\small
\hypersetup{allcolors=black}
\newcommand{\tcarry}[1]{#1}
\newcommand{\tread}[1]{#1}
\newcommand{\tcorr}[1]{#1}
\setlength{\tabcolsep}{3pt}
\renewcommand{\arraystretch}{1.10}
\renewcommand{\tabularxcolumn}[1]{m{#1}}
\begin{tabularx}{\linewidth}{@{}>{\raggedright\arraybackslash}m{.355\linewidth}>{\centering\arraybackslash}m{.30\linewidth}>{\centering\arraybackslash}m{.105\linewidth}>{\centering\arraybackslash}X@{}}
\toprule
\multicolumn{4}{@{}l@{}}{\textit{Common interface:}\quad
 $\bm M_{s+1}=\bm M_s\bm W_s+\vv_s\bm b_s^\top,\qquad
  \bm h_{s+1}=\bm M_{s+1}\bm r_s.$}\\[3pt]
\textbf{Method} & \textbf{Transport $\bm W_s$}
 & \textbf{Write $\bm b_s$} & \textbf{Read $\bm r_s$}\\
\midrule
\textbf{Addition}~\citep{he2016deep} (Standard, Plain Loop)
& $1$ & $1$ & $1$\\[3pt]
\midrule
\multicolumn{4}{@{}l}{\textit{Explicit-history attention}}\\[3pt]
\multicolumn{4}{@{}>{\raggedright\arraybackslash}p{\linewidth}@{}}{History banks start from $\bm M_0=\bm h_0$ and store each new value separately.
At block granularity, the value is $\vd_s$; $\bm e_{\mathrm{new}}$ selects its new storage location.}\\[3pt]
\textbf{AttnRes}~\citep{team2026attention}
& $[\vI,\bm0]$ & $\bm e_{\mathrm{new}}$
& $\bm\pi_{s+1}$\\[1.5pt]
\textbf{MHAR}~\citep{luo2026multi}
& $[\vI,\bm0]$ & $\bm e_{\mathrm{new}}$
& headwise $\bm\pi^{(h)}_{s+1}$\\[3pt]
\multicolumn{4}{@{}l}{\textit{Linear-attention family and related recurrent models}}\\[3pt]
\multicolumn{4}{@{}>{\raggedright\arraybackslash}p{\linewidth}@{}}{The value memory starts from $\bm M_0=0$.
Keys $\vk_s$ and queries $\vq_s$ include each rule's feature mapping, when present.}\\[3pt]
\textbf{Linear (normalized)}~\citep{katharopoulos2020transformers}
& $\vI$ & $\tread{\vk_s}$ & $\displaystyle\tread{\frac{\vq_s}{\bm z_{s+1}^\top\vq_s}}$\\[3pt]
\textbf{RetNet}~\citep{sun2023retentive}
& $\tcarry{\gamma}\vI$ & $\tread{\vk_s}$ & $\tread{\vq_s}$\\[1.5pt]
\textbf{GLA}~\citep{yang2023gated}
& $\tcarry{\Diag(\bm g_s)}$ & $\tread{\vk_s}$ & $\tread{\vq_s}$\\[1.5pt]
\textbf{Mamba-2}~\citep{dao2024transformers}
& $\tcarry{\gamma_s}\vI$ & $\tread{\tau_s\vk_s}$ & $\tread{\vq_s}$\\[1.5pt]
\textbf{HGRN-2}~\citep{qin2024hgrn2}
& $\tcarry{\Diag(\bm g_s)}$ & $\tread{\bm1-\bm g_s}$ & $\tread{\vq_s}$\\[1.5pt]
\textbf{DeltaNet}~\citep{yang2024parallelizing}
& $\vI-\tcorr{\eta_s\vk_s\vk_s^\top}$ & $\tread{\eta_s\vk_s}$ & $\tread{\vq_s}$\\[1.5pt]
\textbf{Gated DeltaNet}~\citep{yang2025gated}
& $\tcarry{\gamma_s}(\vI-\tcorr{\eta_s\vk_s\vk_s^\top})$ & $\tread{\eta_s\vk_s}$ & $\tread{\vq_s}$\\[1.5pt]
\textbf{Kimi Delta Attention (KDA)}~\citep{team2025kimi}
& $\tcarry{\Diag(\bm g_s)}(\vI-\tcorr{\eta_s\vk_s\vk_s^\top})$ & $\tread{\eta_s\vk_s}$ & $\tread{\vq_s}$\\[1.5pt]
\textbf{RWKV-7}~\citep{peng2025rwkv}
& $\tcarry{\Diag(\bm g_s)}-\tcorr{\bm\kappa_s(\bm a_s\odot\bm\kappa_s)^\top}$
& $\tread{\vk_s}$ & $\tread{\vq_s}$\\[3pt]
\textbf{Mesa}~\citep{von2023uncovering}
& $\vI-\vk_s\bm\ell_s^\top$ & $\bm\ell_s$ & $\vq_s$\\[1.5pt]
\multicolumn{4}{@{}>{\raggedright\arraybackslash}p{\linewidth}@{}}{The regression gain $\bm\ell_s$ uses an additional inverse-covariance state.}\\[3pt]
\textbf{LaCT-inspired}~\citep{zhang2026test}
& \multicolumn{2}{c}{Augmented prefix operators (\cref{eq:lact-augmented-map})}
& $[\vq_s;\bm0]$\\[3pt]
\midrule
\multicolumn{4}{@{}l}{\textit{Hyper-connection family:} $\bm M_0=\bm h_0\bm1^\top$; read gives the next input $\bm x_{s+1}$.}\\[3pt]
\textbf{HC}~\citep{zhu2025hyper}
& Learned $\bm W_s$ & Learned $\bm b_s$ & $\tread{\bm p_{s+1}}$\\[3pt]
\textbf{mHC}~\citep{xie2025mhc}
& $\tcarry{\operatorname{Sinkhorn}(\bm U_s)}$
& $\tread{\bm b_s}\ge0$ & $\tread{\bm p_{s+1}}\ge0$\\[3pt]
\textbf{Hyperloop}~\citep{zeitoun2026hyperloop}
& $\tcarry{\Diag(\bm g_s)}$
& $\tread{\bm b_s}\ge0$ & $\tread{\bm p_{s+1}}\ge0$\\[3pt]
\textbf{iHC}\textsuperscript{\ref{fn:ihc-source}}
& $\vI$ & $\tread{\bm b_s}\ge0$ & $\tread{\bm p_{s+1}}\ge0$\\[3pt]
\bottomrule
\end{tabularx}
\endgroup
\end{table}
% Place the framework figure and family table before the Loopback section, so its
% heading, method figure, and equations stay together.
\FloatBarrier

\section{Loopback Transformer}
\label{sec:instance}
\label{sec:design}

\begin{figure}[!htbp]
\centering
\includegraphics[width=\linewidth]{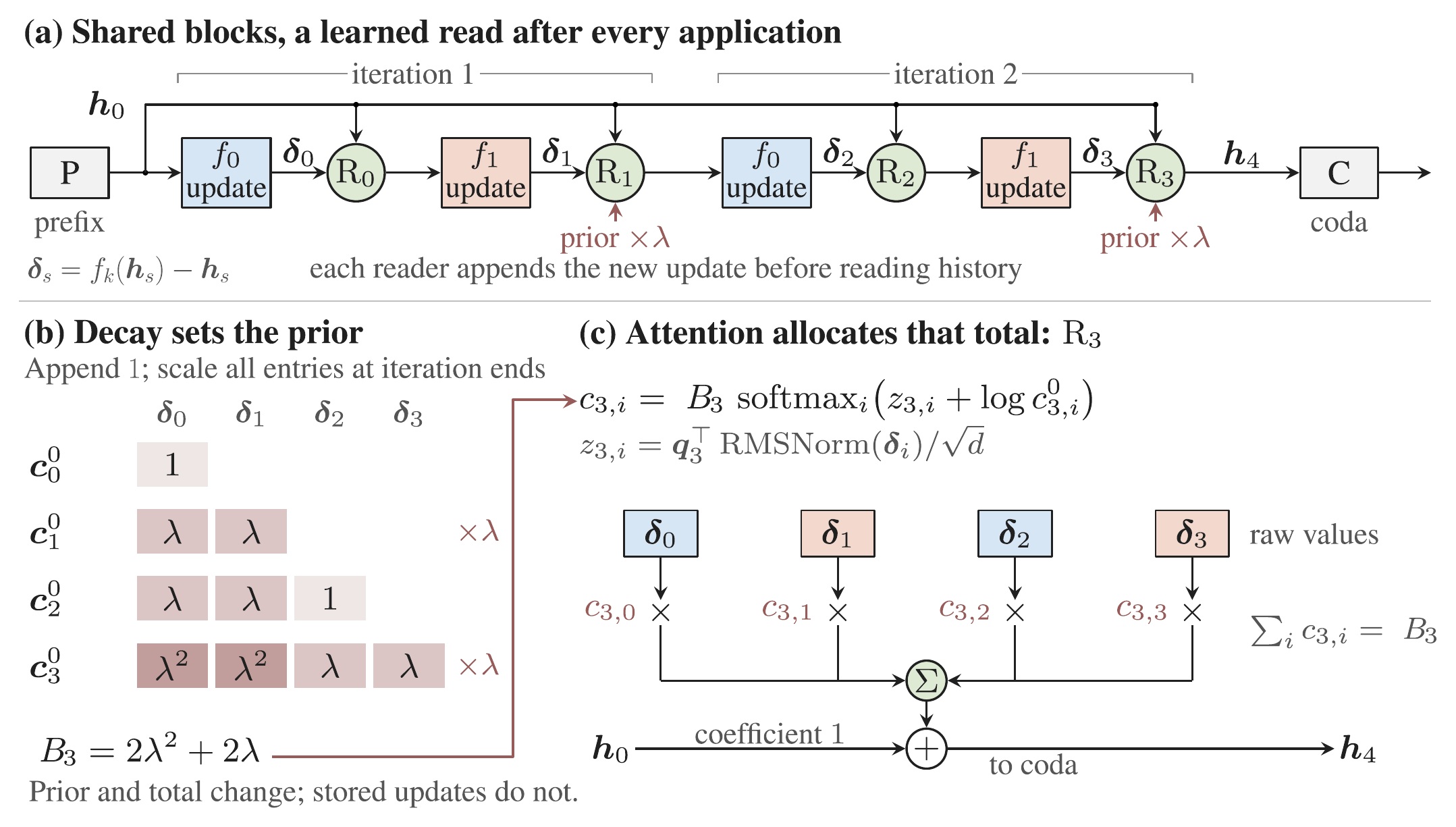}
\caption{\textbf{Loopback separates input, decay, and allocation.}
(a) Shared blocks with causal history reads. (b) Iteration-end decay sets the prior
and total weight. (c) Attention redistributes that total over block updates,
keeping the input coefficient at one.}
\label{fig:arma-method}
\end{figure}

Storing block updates separately allows each read weight to act directly
on one update. Loopback uses this representation to form each new hidden
state from the original loop input $\bm h_0$ and a weighted sum of past
block updates (\cref{fig:arma-method}). Geometric decay bounds
their total weight as loop iterations increase, while attention adjusts
their relative weights using update content. The input stays outside
this allocation with coefficient one, preserving the usual residual
identity path: if all updates vanish, the shared stack returns $\bm h_0$.
Only the residual aggregation changes; the shared blocks are unchanged.

% Approved vector artwork from the refined Figure 3 presentation.

\subsection{Geometric decay sets the prior and total weight}
\label{sec:outer-decay}

High representation similarity has been observed across depth in
Pre-Norm Transformers and across iterations in looped Transformers
\citep{zhu2025hyper,zeitoun2026hyperloop}. If successive block
updates share a common component, ordinary residual addition reinforces
that component at every iteration. Decay limits this accumulation by
shrinking the weights of earlier updates.
The decay-only reference accumulates updates within each loop iteration
and applies a fixed retention $0<\lambda\le1$ at its end.
Each new update enters with coefficient one; at an iteration boundary,
all update coefficients are multiplied by $\lambda$, while the input
coefficient stays at one. For $K=2$, the coefficients of the updates
written so far, in application order, evolve as
\[
    [1]\ \longrightarrow\ [\lambda,\lambda]\ \longrightarrow\
    [\lambda,\lambda,1]\ \longrightarrow\
    [\lambda^2,\lambda^2,\lambda,\lambda].
\]
After two iterations, the first iteration's updates have crossed two
boundaries and the second iteration's updates only one. Repeated scaling
therefore produces geometric weights. In this decay-only reference,
an iteration maps its incoming state $\bm h$ to
\begin{equation}
    \bm h\leftarrow\bm h_0+\lambda\bigl(F(\bm h)-\bm h_0\bigr),
    \qquad F=f_{K-1}\circ\cdots\circ f_0,\quad 0<\lambda\le1.
    \label{eq:anchor}
\end{equation}
Decay follows every iteration, including the last before the coda.
This reference defines Loopback's prior and its trajectory at initialization;
the learned reader below subsequently reweights updates after each block.
No separate reference forward pass is evaluated.

This rule is a scalar instance of the recurrent memory in \cref{eq:memory}.
In the reference, write $\bm h_s=\bm h_0+\bm u_s$, with $\bm u_0=0$.
The accumulated update $\bm u_s$ and the prior coefficient $c^0_{s,i}$
of update $i$ after application $s$ follow
\begin{equation}
    \bm u_{s+1}=\gamma_s\bm u_s+\alpha_s\vd_s,
    \qquad c^0_{s,i}=\alpha_i\prod_{j=i+1}^{s}\gamma_j.
    \label{eq:scalar-retention}
\end{equation}
Here $\alpha_s$ scales the new write and $\gamma_s$ retains the previous
accumulation. Both equal $\lambda$ when $(s+1)\bmod K=0$ and one otherwise.
Each write is scaled on entry
by $\alpha_i$ and subsequently by the retention factors $\gamma_j$.
For $0\le i\le s<KR$, counting iteration boundaries from write $i$
through read $s$ gives
\begin{equation}
    c^0_{s,i}=\lambda^{\lfloor(s+1)/K\rfloor-\lfloor i/K\rfloor},
    \qquad B_s=\sum_{i=0}^{s}c^0_{s,i}.
    \label{eq:boundary-schedule}
\end{equation}
Superscript $0$ marks the prior, and $B_s$ is its total weight at read $s$.
For $\lambda<1$, the prior favors later iterations. After $R$ complete
iterations, $B_{KR-1}=K\sum_{j=1}^{R}\lambda^j$, which equals $KR$
without decay and is bounded by
$K\lambda/(1-\lambda)$ for fixed $0<\lambda<1$.
Thus a fixed retention controls accumulation as the number of iterations grows.

\subsection{Softmax attention reweights the updates}
\label{sec:inner-allocation}

The geometric prior assigns weights by loop iteration. Earlier updates
may remain useful for forming the next hidden state, and updates from
the same iteration can play different roles despite receiving equal
prior weights. We therefore use attention to adjust their weights
using each token's update content while preserving the total set by decay.
Loopback is single-stream, so $\bm x_s=\bm h_s$. After application $s$,
we append $\vd_s=f_k(\bm h_s)-\bm h_s$, with $k=s\bmod K$, to its history.
These updates come from the current Loopback trajectory; only the prior
coefficients are fixed by the schedule above.
For the read after this application, a learned vector $\vq_s\in\R^d$
scores normalized updates $\vk_i$ for $0\le i\le s$.
Using $\varepsilon>0$ for numerical stability, the scores $z_{s,i}$,
weights $c_{s,i}$, and next state are
\begin{align}
    \vk_i&=\frac{\vd_i}{\sqrt{d^{-1}\|\vd_i\|^2+\varepsilon}},
    \qquad z_{s,i}=\vq_s^\top\vk_i/\sqrt d,\notag\\
    c_{s,i}&=B_s\operatorname{softmax}_i(z_{s,i}+\log c^0_{s,i})
       =B_s\frac{c^0_{s,i}e^{z_{s,i}}}
                    {\sum_{j=0}^{s}c^0_{s,j}e^{z_{s,j}}},
    \quad \bm h_{s+1}=\bm h_0+\sum_{i=0}^{s}c_{s,i}\vd_i .
    \label{eq:instance}
\end{align}
Each baseline weight $c^0_{s,i}$ is multiplied by the content factor
$e^{z_{s,i}}$; normalization then restores the total $B_s$. Attention can
therefore favor an older update while preserving $\sum_i c_{s,i}=B_s$.
This total depends on depth but is unchanged by attention at each read.
The query is specific to application $s$ and shared across tokens;
content enters through the keys. There are no query, key, or value
projections. The language-model loss trains the queries to redistribute weight
(\cref{app:allocation-gradient}).
The readout replaces the current mixture of updates and retains the input
coefficient of one. It feeds the next block, or the coda after the final
application; weighting does not modify the stored updates.
We initialize the queries to zero, giving $c_{s,i}=c^0_{s,i}$ and recovering
the decay-only trajectory. With zero queries, $\lambda=1$ also recovers
ordinary addition. The first read has only one update and needs no learned query
(\cref{app:implementation}).

Loopback applies the write--transport--read view of linear attention and
recurrent memory along depth: blocks supply the values, retention defines
their prior weights,
and attention reads them to form later block inputs.
Its explicit history allows individual updates to be reweighted.
The bank requires $O(KRd)$ storage, and total
read work is $O((KR)^2d)$ per token beyond block computation.

\section{Experiments}
\label{sec:experiments}
\begingroup
\setlength{\intextsep}{6pt plus 2pt minus 2pt}

We evaluate parameter efficiency, residual designs, and readers on
ClimbMix \citep{diao2026nemotron}. Main comparisons match training tokens and
executed depth within each scale.
Scale names and the 20-tokens-per-parameter (TPP) budget use Standard's total
parameters; Transformer counts exclude embeddings and the
vocabulary head. Training and evaluation details are in \cref{app:190m}.

% The paper's main table (tab:downstream-main): all nine rules at 191M (main_191m.json) and in the depth sweep at
% 374M with 32 layers and 778M with 48 layers (depth_sweep_data.json, transformer_params.json); thirteen zero-shot
% tasks with the beginning-of-sequence token prepended to every prompt. Generated by
% figure/depth_sweep/make_depth_assets.py (outside this repository); regenerate instead of editing by hand.
% Floats ([!tp]): at one page tall it needs its own place.
\begin{table}[!tp]
\setlength{\abovecaptionskip}{4pt}
\caption{\textbf{Residual adapters at matched tokens and depth.}
Validation loss, zero-shot accuracy (\%), and WikiText perplexity at 191M (20 layers), 374M (32 layers), and
778M (48 layers) (\cref{tab:depth-shapes}); every prompt starts with the beginning-of-sequence token, as
in training (\cref{app:190m-evaluation}). Arrows mark lower-is-better metrics; bold marks the best result
per row. Percentages compare Transformer parameters with Standard, excluding embeddings and the vocabulary
head and counting shared weights once.}
\label{tab:downstream-main}
\centering\footnotesize
\setlength{\tabcolsep}{2.3pt}
\renewcommand{\arraystretch}{0.94}
\begin{tabular*}{\linewidth}{@{\extracolsep{\fill}}lrrrrrrrr>{\columncolor{paperTeal!7}}r@{}}
\toprule
& & & \multicolumn{4}{c}{Hyper-connection family} & \multicolumn{2}{c}{Attention family} & \\
\cmidrule(lr){4-7}\cmidrule(lr){8-9}
Benchmark & Standard & \shortstack{Plain\\Loop} & mHC & HC & iHC & Hyperloop & MHAR & AttnRes & \textbf{Loopback} \\
\midrule
\multicolumn{10}{@{}l}{\color{paperNavy}\textbf{191M, 20 layers, $\looptopology{2}{4}{4}{2}$, 40.0\% of Standard, 3.82B tokens (20~TPP)}} \\
\midrule
Transformer params (M) & 141.59 & 56.64 & 56.93 & 56.65 & 57.04 & 56.79 & 56.64 & 56.66 & 56.65 \\
\textbf{ClimbMix loss} $\downarrow$ & 2.6234 & 2.6802 & 2.6250 & 2.6449 & 2.6564 & 2.6489 & 2.6564 & 2.6404 & \textbf{2.6187} \\
\addlinespace[0.5pt]
\quad PIQA & 65.51 & 64.96 & \textbf{65.89} & 64.74 & 65.34 & 65.07 & 65.13 & 64.47 & 65.78 \\
\quad HellaSwag & 33.26 & 32.14 & 33.79 & 33.18 & 32.74 & 33.04 & 33.12 & 33.07 & \textbf{33.83} \\
\quad WinoGrande & 51.70 & 51.38 & 52.64 & 51.22 & 51.70 & 49.80 & 50.75 & \textbf{52.72} & 51.14 \\
\quad OpenBookQA & 32.60 & 29.40 & 31.40 & 30.20 & \textbf{32.80} & 31.80 & 31.80 & 29.20 & 32.40 \\
\quad CommonsenseQA & 21.54 & 19.98 & 19.90 & 18.67 & 19.98 & \textbf{21.95} & 18.92 & 20.48 & 21.13 \\
\addlinespace[0.5pt]
\quad BoolQ & \textbf{62.72} & 46.82 & 61.99 & 59.60 & 62.17 & 60.58 & 61.22 & 60.73 & 61.44 \\
\quad RACE & 28.04 & 29.09 & 27.75 & 29.86 & 29.67 & 28.61 & \textbf{30.14} & 29.95 & 28.33 \\
\addlinespace[0.5pt]
\quad ARC-Easy & \textbf{47.90} & 46.93 & 46.51 & 47.14 & 46.42 & 45.66 & 46.21 & 44.28 & 47.22 \\
\quad ARC-Challenge & 26.45 & 25.94 & 25.68 & 25.94 & 25.09 & 26.11 & \textbf{26.62} & 26.11 & 25.77 \\
\addlinespace[0.5pt]
\quad MMLU (57 subjects) & 23.56 & 24.57 & 24.01 & 24.13 & 23.64 & 24.02 & 23.25 & \textbf{26.57} & 23.68 \\
\addlinespace[0.5pt]
\quad ANLI-R1 & 30.40 & 32.20 & 27.30 & 32.40 & 31.30 & 32.90 & \textbf{33.00} & 31.60 & 30.70 \\
\quad RTE & 49.46 & 50.90 & \textbf{55.60} & 53.43 & 53.43 & 53.79 & 53.79 & 53.07 & 54.15 \\
\addlinespace[0.5pt]
\quad LAMBADA & 24.47 & 22.01 & 25.46 & 25.38 & 24.92 & 24.72 & 23.91 & 24.18 & \textbf{26.80} \\
\quad WikiText word ppl $\downarrow$ & 46.30 & 50.30 & 45.86 & 47.52 & 48.72 & 47.87 & 48.30 & 47.01 & \textbf{45.74} \\
\addlinespace[0.5pt]
\textbf{Mean} (13 tasks) & 38.28 & 36.64 & 38.30 & 38.15 & 38.40 & 38.31 & 38.30 & 38.19 & \textbf{38.64} \\
\addlinespace[2pt]\midrule
\multicolumn{10}{@{}l}{\color{paperNavy}\textbf{374M, 32 layers, $\looptopology{2}{14}{2}{2}$, 56.3\% of Standard, 7.47B tokens (20~TPP)}} \\
\midrule
Transformer params (M) & 323.02 & 181.70 & 181.87 & 181.71 & 182.52 & 181.79 & 181.70 & 181.75 & 181.72 \\
\textbf{ClimbMix loss} $\downarrow$ & 2.4395 & 2.4371 & 2.4118 & 2.4273 & 2.4199 & 2.4147 & 2.4310 & 2.4166 & \textbf{2.4008} \\
\addlinespace[0.5pt]
\quad PIQA & 68.82 & 68.23 & 68.39 & 68.88 & \textbf{69.48} & 69.21 & 69.10 & 68.72 & 69.10 \\
\quad HellaSwag & 38.72 & 39.47 & 40.44 & 39.89 & 40.36 & 40.48 & 39.33 & 40.42 & \textbf{41.11} \\
\quad WinoGrande & 51.93 & \textbf{52.96} & 52.88 & 52.17 & 52.01 & 52.49 & 52.88 & 51.14 & 52.09 \\
\quad OpenBookQA & 30.80 & 34.00 & 33.60 & 32.80 & 31.20 & 32.00 & 33.60 & 33.20 & \textbf{34.60} \\
\quad CommonsenseQA & 19.98 & 20.23 & 19.90 & 21.05 & 20.31 & \textbf{21.21} & 20.64 & \textbf{21.21} & 19.74 \\
\addlinespace[0.5pt]
\quad BoolQ & 47.13 & 60.89 & 58.04 & 58.65 & \textbf{62.42} & 56.73 & 55.99 & 58.23 & 58.90 \\
\quad RACE & 29.28 & \textbf{31.00} & 30.81 & 29.38 & 30.14 & 28.42 & 30.91 & 30.81 & 29.67 \\
\addlinespace[0.5pt]
\quad ARC-Easy & 51.26 & 52.90 & 53.37 & 51.43 & 52.65 & \textbf{54.63} & 52.44 & 51.73 & 52.15 \\
\quad ARC-Challenge & 27.65 & 28.58 & 29.01 & 28.41 & 27.82 & 28.41 & \textbf{29.69} & 28.16 & 28.07 \\
\addlinespace[0.5pt]
\quad MMLU (57 subjects) & 25.25 & 23.64 & 25.32 & 25.76 & 24.85 & 25.81 & \textbf{26.25} & 25.37 & 26.24 \\
\addlinespace[0.5pt]
\quad ANLI-R1 & 33.30 & 33.60 & 32.00 & 30.70 & 32.60 & 31.30 & 33.40 & \textbf{33.90} & 32.40 \\
\quad RTE & \textbf{54.87} & 53.79 & 53.43 & 51.26 & 51.99 & 50.18 & 53.07 & 46.57 & 53.07 \\
\addlinespace[0.5pt]
\quad LAMBADA & 30.29 & 31.69 & 31.94 & 31.85 & 32.54 & 32.76 & 31.19 & 33.05 & \textbf{34.43} \\
\quad WikiText word ppl $\downarrow$ & 34.79 & 34.57 & 33.34 & 34.05 & 33.69 & 33.45 & 34.23 & 33.59 & \textbf{32.61} \\
\addlinespace[0.5pt]
\textbf{Mean} (13 tasks) & 39.18 & 40.84 & 40.70 & 40.17 & 40.64 & 40.28 & 40.65 & 40.19 & \textbf{40.89} \\
\addlinespace[2pt]\midrule
\multicolumn{10}{@{}l}{\color{paperNavy}\textbf{778M, 48 layers, $\looptopology{3}{21}{2}{3}$, 56.3\% of Standard, 15.60B tokens (20~TPP)}} \\
\midrule
Transformer params (M) & 708.68 & 398.63 & 398.84 & 398.65 & 400.11 & 398.74 & 398.64 & 398.73 & 398.68 \\
\textbf{ClimbMix loss} $\downarrow$ & 2.2644 & 2.2737 & 2.2454 & 2.2642 & 2.2523 & 2.2539 & 2.2646 & 2.2529 & \textbf{2.2404} \\
\addlinespace[0.5pt]
\quad PIQA & 71.44 & 71.55 & \textbf{72.20} & 71.76 & 70.57 & 71.16 & 71.49 & 71.71 & 71.16 \\
\quad HellaSwag & 47.07 & 47.07 & \textbf{49.03} & 47.74 & 48.56 & 48.08 & 48.10 & 48.72 & 48.97 \\
\quad WinoGrande & 52.09 & 49.80 & 53.43 & 53.04 & 53.12 & 53.28 & 53.51 & \textbf{55.64} & 53.91 \\
\quad OpenBookQA & 34.80 & 35.40 & 35.20 & 35.00 & 35.60 & 36.00 & 34.00 & \textbf{36.40} & 36.20 \\
\quad CommonsenseQA & 19.16 & 19.41 & 18.10 & 19.82 & 18.76 & 20.97 & 20.15 & \textbf{21.95} & 20.80 \\
\addlinespace[0.5pt]
\quad BoolQ & 59.97 & 58.99 & 55.17 & 60.28 & 58.93 & \textbf{60.37} & 57.98 & 57.92 & 58.93 \\
\quad RACE & 31.48 & 32.54 & 31.96 & 32.06 & 32.34 & 32.44 & \textbf{33.49} & 33.21 & 33.30 \\
\addlinespace[0.5pt]
\quad ARC-Easy & 60.44 & 57.58 & 61.20 & 60.56 & 60.98 & 58.63 & \textbf{61.41} & 58.42 & 58.92 \\
\quad ARC-Challenge & 31.91 & 31.40 & 31.83 & 32.25 & \textbf{33.19} & 31.06 & 32.59 & 31.74 & 32.08 \\
\addlinespace[0.5pt]
\quad MMLU (57 subjects) & \textbf{25.77} & 25.10 & 25.70 & 25.37 & 24.53 & 24.84 & 25.49 & 25.44 & 25.30 \\
\addlinespace[0.5pt]
\quad ANLI-R1 & 31.30 & 32.00 & 33.70 & 33.20 & 30.70 & 33.30 & 33.20 & 33.30 & \textbf{35.20} \\
\quad RTE & 56.32 & \textbf{57.40} & 52.71 & 51.26 & 53.79 & 53.79 & 50.18 & 53.07 & 51.99 \\
\addlinespace[0.5pt]
\quad LAMBADA & 37.76 & 37.84 & 39.71 & 38.87 & \textbf{39.78} & 38.07 & 38.54 & 37.92 & 39.74 \\
\quad WikiText word ppl $\downarrow$ & 26.39 & 26.78 & 25.68 & 26.49 & 25.84 & 26.01 & 26.35 & 25.95 & \textbf{25.33} \\
\addlinespace[0.5pt]
\textbf{Mean} (13 tasks) & 43.04 & 42.78 & 43.07 & 43.17 & 43.14 & 43.23 & 43.09 & 43.50 & \textbf{43.58} \\
\bottomrule
\end{tabular*}
\end{table}

\paragraph{Baselines.}
The standard Transformer uses independent blocks; Plain Loop repeats shared blocks with
ordinary addition. HC learns full stream mixing, mHC uses Sinkhorn-normalized
transport, Hyperloop uses diagonal transport, and iHC uses identity transport
with learned read and write gates (\cref{tab:residual-families}). AttnRes and
MHAR are our loop-level adaptations of Attention Residuals and multi-head attention residuals
\citep{team2026attention,luo2026multi}. Adapters act only inside the loop.
Looped arms share the topology and learning rate within each setting;
adapter placement and memory sources differ.

\subsection{Parameter-efficient language modeling}
\label{sec:compression}

We first test whether Loopback can achieve lower validation loss than depth-matched
standard Transformers with fewer Transformer parameters. \Cref{tab:downstream-main}
compares all nine rules at 191M with 20 layers, 374M with 32 layers, and 778M with
48 layers. Loopback stores 40.0\%, 56.3\%, and 56.3\% of the standard
baseline's Transformer parameters, respectively, and attains the lowest validation
loss at all three. Its losses are 0.005, 0.039, and 0.024 nats lower than those of
the standard Transformer, respectively.
mHC is the closest looped baseline at all three, trailing Loopback by 0.006, 0.011,
and 0.005 nats. At 374M with 24 layers, Loopback stores 58.3\% of the standard
baseline's Transformer parameters and again has the lowest loss among the nine
rules (\cref{tab:depth-loss}). AttnRes trails Loopback by 0.022 nats
at 191M and 0.009 nats at 374M with 24 layers.
\Cref{fig:train-loss} shows the training-loss curves of Loopback and
the standard Transformer across all five model scales.

% Figure: training-loss curves at the opening of Section 6. Generated by figure/train_loss/fig_train_loss_5.py (outside
% this repository; it reuses fig_train_loss.py's style and maps the Dense data key to the Standard label) from the runs'
% training logs (train_loss_data.json there, and b3n_train_loss.json for 3B). At 374M and 778M each curve follows the
% 5-TPP rung to its pre-decay fork and the 20-TPP rung after it (one uninterrupted trajectory); at 1.4B (N4-3 runs of
% 2026-10-03, dense25 and [3->8x2->6]) and 3B (N4-3, dense24 and [6->6x2->6], 2026-10-22/25) the 2-, 5- and 20-TPP
% rungs are stitched the same way. Regenerate fig_train_loss.pdf after any change to those inputs.
% Keep text eligible below the curves and main table; avoid a float-only page.
\begin{figure}[!htbp]
\centering
\includegraphics[width=\linewidth]{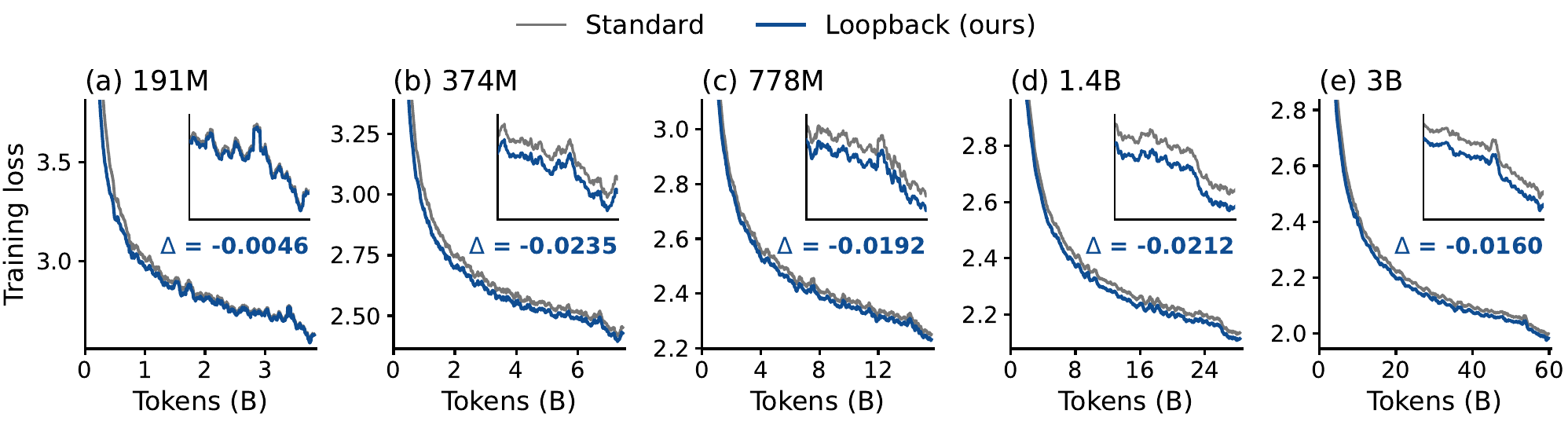}
\caption{\textbf{Training loss of the standard Transformer and Loopback.} 100-step moving averages against tokens seen; insets
show the last quarter of training; $\Delta$ is the final training-loss difference, Loopback minus the standard Transformer, over the
last 100 steps. The five panels use 20, 24, 24, 25, and 24 executed layers, respectively.
Loopback stores 40--75\% of the standard baseline's Transformer parameters.}
\label{fig:train-loss}
\end{figure}

Loopback also has the highest thirteen-task mean and the lowest WikiText
perplexity in all three settings of \cref{tab:downstream-main}, although
individual-task results remain mixed. Across the tested training seeds,
Loopback retains lower validation loss than the looped baselines retrained
at 191M and than the standard Transformer and mHC at 374M with 24 layers
(\cref{app:seeds}).

\Needspace{4\baselineskip}
\subsection{Parameter efficiency at larger scale}
\label{sec:scale-760m}

At 1.4B and 3B, Loopback uses $\looptopology{3}{8}{2}{6}$ and
$\looptopology{6}{6}{2}{6}$, respectively, matching the standard Transformer's
25 and 24 executed layers with 68\% and 75\% of its Transformer parameters.
We evaluate these 20-TPP checkpoints on general benchmarks and mathematical
and code tasks. Tests of additional loop iterations also include 191M and
374M models for comparison with Plain Loop.

\subsubsection{General benchmarks}
\label{sec:large-general}

\Cref{fig:scale-large-general} summarizes Transformer parameter counts, validation loss,
and the thirteen-task zero-shot mean. Loopback improves both aggregate metrics
at each scale, and has lower WikiText perplexity: 20.29 versus 21.09 at 1.4B
and 16.74 versus 17.28 at 3B. Individual tasks are mixed, with Loopback ahead
on nine and eleven of the thirteen tasks, respectively; all task scores are
in \cref{tab:scale-large}.
\begin{figure}[!htbp]
\centering
\includegraphics[width=\linewidth]{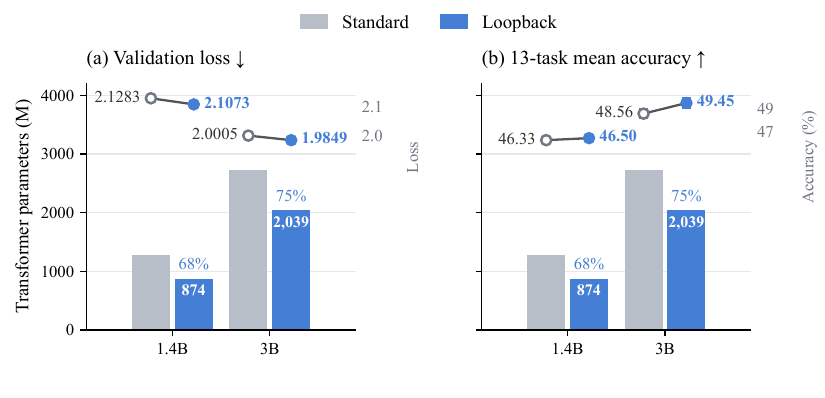}
\caption{\textbf{General benchmarks at 1.4B and 3B.} Loopback uses 68\% and 75\%
of Standard's Transformer parameters while attaining lower validation loss and higher
thirteen-task mean accuracy. Bars use the left parameter axis; connected points
use the right metric axis. Training tokens and executed depth are matched
within each scale, with one run per arm at 20~TPP.}
\label{fig:scale-large-general}
\end{figure}

\Needspace{4\baselineskip}
\subsubsection{Code and mathematical evaluation}
\label{sec:large-math}

We test whether shared computation improves the modeling of mathematical
and code solutions with fewer Transformer parameters. At matched
training tokens and executed depth, reference-solution loss on GSM8K,
MATH-500, HumanEval, and MBPP measures how well the model predicts a worked
solution given the problem. Five-shot numeric-answer accuracy on six
word-problem sets separately tests final-answer correctness.
Loopback has lower solution loss on all four datasets at both large scales
(\cref{tab:scale-large-math}). At 3B, word-problem accuracy rises from 2.6\%
to 6.4\%. The nine-design comparisons
at 191M and 374M also give Loopback the lowest four-set mean (\cref{tab:mathreason}). Evaluation protocols are in \cref{app:mathreason}.

% Generated by figures/make_scale_large_extensions.py from the JSON and CSV inputs.
\begin{table}[!htbp]
\setlength{\abovecaptionskip}{4pt}
\caption{\textbf{Mathematical and code evaluation at 1.4B and 3B.}
Reference-solution loss is teacher-forced NLL (nats per solution token), not generation
or code-execution accuracy. Word problems use mean five-shot numeric-answer accuracy over six sets.
$\Delta$ is Loopback minus Standard (nats for loss; percentage points for accuracy).
Training tokens and depth are matched, with one run per arm. Shading marks Loopback;
bold scores are row-best, and green differences favor Loopback.}
\label{tab:scale-large-math}
\centering\footnotesize
\setlength{\tabcolsep}{3pt}
\renewcommand{\arraystretch}{1.08}
\begin{tabular*}{\linewidth}{@{\extracolsep{\fill}}lr>{\columncolor{paperTeal!7}}rrr>{\columncolor{paperTeal!7}}rr@{}}
\toprule
 & \multicolumn{3}{c}{1.4B} & \multicolumn{3}{c}{3B} \\
\cmidrule(lr){2-4}\cmidrule(l){5-7}
Reference-solution loss $\downarrow$ & Standard & \textbf{Loopback} & $\Delta$ & Standard & \textbf{Loopback} & $\Delta$ \\
\midrule
GSM8K & 1.275 & \textbf{1.227} & \textcolor{paperGreen}{-0.048} & 1.143 & \textbf{1.086} & \textcolor{paperGreen}{-0.057} \\
MATH-500 & 1.421 & \textbf{1.391} & \textcolor{paperGreen}{-0.030} & 1.289 & \textbf{1.259} & \textcolor{paperGreen}{-0.030} \\
HumanEval & 1.065 & \textbf{1.035} & \textcolor{paperGreen}{-0.030} & 0.906 & \textbf{0.868} & \textcolor{paperGreen}{-0.038} \\
MBPP & 1.282 & \textbf{1.235} & \textcolor{paperGreen}{-0.047} & 1.104 & \textbf{1.078} & \textcolor{paperGreen}{-0.026} \\
\midrule
\textbf{Four-set mean} & 1.261 & \textbf{1.222} & \textcolor{paperGreen}{-0.039} & 1.110 & \textbf{1.073} & \textcolor{paperGreen}{-0.037} \\
\midrule
Word problems (\%) $\uparrow$ & 2.0 & \textbf{2.6} & \textcolor{paperGreen}{+0.6} & 2.6 & \textbf{6.4} & \textcolor{paperGreen}{+3.8} \\
\bottomrule
\end{tabular*}
\end{table}

\subsubsection{Robustness to additional visits}
\label{sec:large-visits}

Shared weights allow extra visits at inference, but excessive recurrence can
degrade predictions, a problem known as overthinking \citep{bansal2022overthinking}.
Ouro and the STARS study report this sensitivity beyond the trained loop
count in language models \citep{zhu2025ouro,yang2026stars}. We test whether
Loopback reduces it without retraining, comparing with Plain Loop's ordinary
residual addition. Extra destinations reuse the last trained visit's queries.

At 191M, 374M, and 1.4B, Loopback's loss rises less across all tested extra
counts, up to twice the trained count (\cref{fig:scale-large-visits}a--c).
At eight visits, the 191M loss increases by 0.060 for Loopback versus 0.454
for Plain Loop; at four visits, the 374M increases are 0.074 versus 0.270,
and the 1.4B increases are 0.028 versus 0.248.
On these curves, Loopback degrades less than Plain Loop when visits are added.
At 3B, one extra iteration raises Loopback's thirteen-task mean from
49.45\% to 50.17\% (\cref{fig:scale-large-visits}d); only Loopback is
evaluated at extra depth, with Standard as a fixed-depth reference.
The thirteen individual 3B task results are in \cref{app:extra-visits}.

\begin{figure}[!htbp]
\centering
\includegraphics[width=\linewidth]{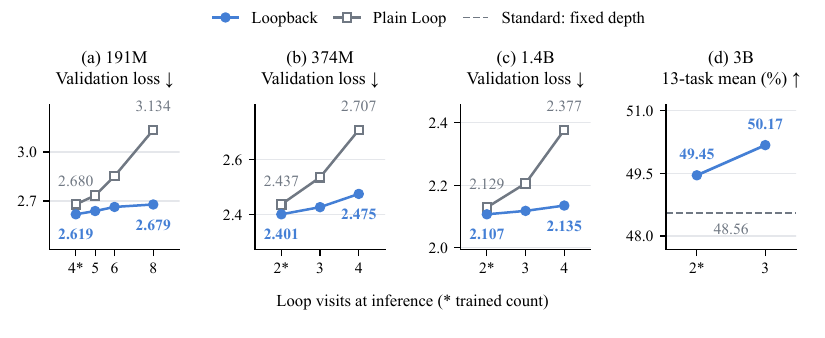}
\caption{\textbf{Robustness to additional visits without retraining.}
(a--c) Validation loss (nats/token), trained at 20, 32, and 25 layers;
(d) thirteen-task mean accuracy, trained at 24 layers.
Asterisks mark trained visit counts; the dashed line shows the fixed-depth
Standard reference at 3B.}
\label{fig:scale-large-visits}
\end{figure}

\Needspace{4\baselineskip}
\subsection{Ablations: decay, attention, and initialization}
\label{sec:ablation}
\label{sec:init}

We test the contributions of decay and attention and whether Loopback's gains
persist across weight initializations. Decay controls accumulated update weight;
attention allocates it by content. \Cref{tab:ablation} tests these components
separately and together, under three initialization schemes:
fixed-scale Gaussian, GPT-2 residual scaling
\citep{radford2019language}, and PyTorch defaults. Hyperloop reports that its
advantage over Standard does not persist after switching to GPT-2-style
initialization and retuning learning rates \citep{zeitoun2026hyperloop}, so we
check Loopback's sensitivity to initialization, holding learning rates fixed.

% Three panels. Shaded cells are Loopback; green = Loopback minus Standard.
% Panels (b--c) reuse the initialization data in tables_init.tex.
\newcommand{\gainvsstandard}[1]{{\scriptsize\color{paperGreen}(#1)}}
% Initialization rows included as panels (b--c) of the main ablation table.
% Numerical entries are unchanged from the former appendix table.
\newcommand{\weightinitializationrows}{%
\multicolumn{5}{@{}l}{\color{paperNavy}\textbf{(b) Weight initialization: 191M, 20 layers}}\\[2pt]
Model & Transformer params (M) & $\mathcal N(0,0.02^2)$ & GPT-2 & PyTorch \\
\midrule
Standard & 141.59 & 2.6234 & 2.6419 & 2.7145 \\
Plain Loop & 56.64 & 2.6802 & 2.6977 & 2.6642 \\
mHC & 56.93 & 2.6250 & 2.6408 & 2.6278 \\
\rowcolor{paperTeal!7}
\textbf{Loopback} & 56.65 & \textbf{2.6187} \gainvsstandard{$-$0.0047} & \textbf{2.6178} \gainvsstandard{$-$0.0241} & \textbf{2.6138} \gainvsstandard{$-$0.1007} \\
\midrule
\multicolumn{5}{@{}l}{\color{paperNavy}\textbf{(c) Weight initialization: 374M, 24 layers}}\\[2pt]
Model & Transformer params (M) & $\mathcal N(0,0.02^2)$ & GPT-2 & PyTorch \\
\midrule
Standard & 308.33 & 2.4198 & 2.4252 & 2.4296 \\
Plain Loop & 179.86 & 2.4415 & 2.4272 & 2.4334 \\
mHC & 180.06 & 2.4062 & \textbf{2.4091} & 2.4093 \\
\rowcolor{paperTeal!7}
\textbf{Loopback} & 179.88 & \textbf{2.3971} \gainvsstandard{$-$0.0227} & 2.4117 \gainvsstandard{$-$0.0135} & \textbf{2.3938} \gainvsstandard{$-$0.0358} \\
}
\begin{table}[!htb]
\caption{\textbf{Decay, attention, and weight initialization.} Validation loss ($\downarrow$)
at 20 layers (191M) and 24 layers (374M);
shading marks Loopback and green gives its difference from Standard.
(a) Fixed allocation uses zero queries.
(b--c) Three weight-initialization schemes with learning rates held fixed;
bold marks the lowest loss in each initialization column. Parameter counts exclude
embeddings and the vocabulary head.}
\label{tab:ablation}
\label{tab:init}
\centering\small
\setlength{\tabcolsep}{6pt}
\renewcommand{\arraystretch}{1.10}
\begin{tabularx}{\linewidth}{@{}p{.24\linewidth}*{4}{>{\centering\arraybackslash}X}@{}}
\toprule
\multicolumn{5}{@{}l}{\color{paperNavy}\textbf{(a) Decay and attention at two scales}}\\[2pt]
& \multicolumn{2}{c}{191M (Standard: 2.6234)} & \multicolumn{2}{c@{}}{374M (Standard: 2.4198)}\\
\cmidrule(lr){2-3}\cmidrule(l){4-5}
Decay & Fixed & Learned & Fixed & Learned\\
\midrule
Off ($\lambda=1$) & 2.6802 & 2.6621 & 2.4415 & 2.4308\\
On ($\lambda=1/4$) & 2.6220 & \cellcolor{paperTeal!7}\textbf{2.6187} \gainvsstandard{$-$0.0047} & 2.3999 & \cellcolor{paperTeal!7}\textbf{2.3971} \gainvsstandard{$-$0.0227}\\
\midrule
\weightinitializationrows
\bottomrule
\end{tabularx}
\end{table}

The complete design gives the lowest loss at both scales in panel (a): removing
attention increases loss by 0.0033 and 0.0028 nats at 191M and 374M, respectively;
removing decay increases it by 0.0434 and 0.0337. Thus content-based allocation
improves on decay's fixed weights, and controlling accumulation also benefits
the learned reader. Across the six
scale--initialization settings in panels (b--c), Loopback outperforms Standard
and Plain Loop throughout and attains the lowest loss in five.

The default Gaussian scheme draws linear and embedding weights from
$\mathcal N(0,0.02^2)$. GPT-2 \citep{radford2019language} additionally scales
the standard deviation of attention-output and MLP down-projections by
$1/\sqrt{2D}$, where $D$ is the executed depth. PyTorch uses Kaiming-uniform
linear weights with $a=\sqrt5$ and $\mathcal N(0,1)$ embeddings.
These choices probe fixed-scale, depth-scaled, and fan-in-scaled linear weights,
with different embedding scales.

\FloatBarrier
\Needspace{4\baselineskip}
\subsection{Depth at fixed parameters}
\label{sec:depth}

We ask whether Loopback's validation-loss advantage persists when a fixed parameter budget
is allocated to different widths and depths, instead of depending on one
chosen topology. At 374M and 778M, we train every configuration from scratch,
holding training tokens fixed and total parameters approximately fixed while
trading width for depth from 16 to 64 layers. Our loop visits stay fixed at two.
We compare the standard Transformer and the eight looped rules at every depth (\cref{fig:depth},
\cref{tab:depth-loss}, \cref{app:depth}). The effect of added depth differs
across residual rules. The standard Transformer loses ground
as it gets deeper and narrower: it is best at 24 layers and, at 64, is 0.076
worse at 374M and 0.035 worse at 778M. Among the looped rules, Loopback is lowest at
nine of the ten points, and its best tested depth moves from 24 layers at 374M to 48 at
778M. At the sweep endpoints, its margin over the standard Transformer increases
from 0.003 at 16 layers to 0.073 at 64 layers at 374M, and from 0.002 to 0.028 at 778M.

% Figure: depth at fixed parameters (Experiments, "Depth at fixed parameters"). Generated by figure/depth_sweep/make_depth_assets.py (outside this
% repository) from figure/depth_sweep/depth_sweep_data.json, the values of tab:depth-loss, at print size (5.5 in wide);
% `make_depth_assets.py --paper <this repository>` regenerates fig_depth_sweep.pdf only (--tables also rewrites the tables).
\begin{figure}[!htb]
\centering
\includegraphics[width=0.96\linewidth]{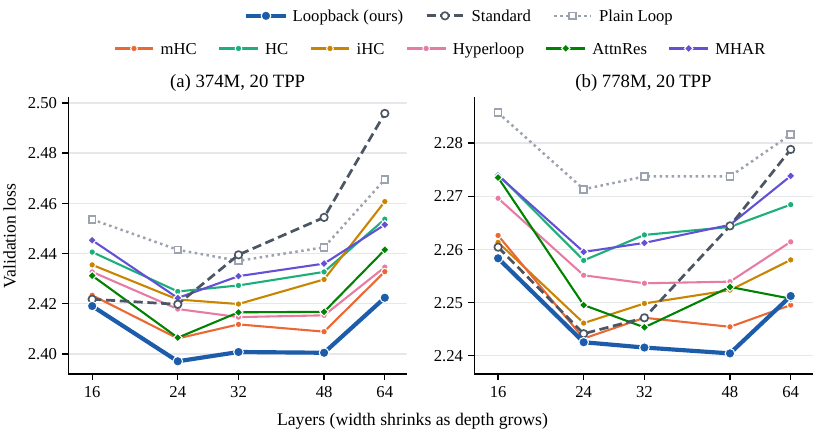}
\caption{\textbf{Depth at fixed parameters.} ClimbMix validation loss after 20~TPP as each scale trades width for
depth at approximately fixed total parameters and fixed training tokens. Looped rules make two visits through a shared
stack that stores 56--58\% of the standard baseline's Transformer parameters.}
\label{fig:depth}
\end{figure}

\Needspace{4\baselineskip}
\subsection{Linear attention and recurrent memory in looped Transformers}
\label{sec:memory-families}

We adapt the nine recurrent cores, Mesa, and a LaCT-inspired reader in
\cref{tab:families} to visit boundaries at 191M and 374M, leaving token attention
unchanged. Each reader uses either cumulative visit states or whole-visit
updates; both interfaces initialize to Plain Loop (\cref{app:readers}).

% Generated by figures/make_memory_readers.py from figures/memory_readers.csv.
% The full losses are in tab:readers-compressed.
\begin{figure}[!htb]
\setlength{\abovecaptionskip}{5pt}
\centering
\includegraphics[width=\linewidth]{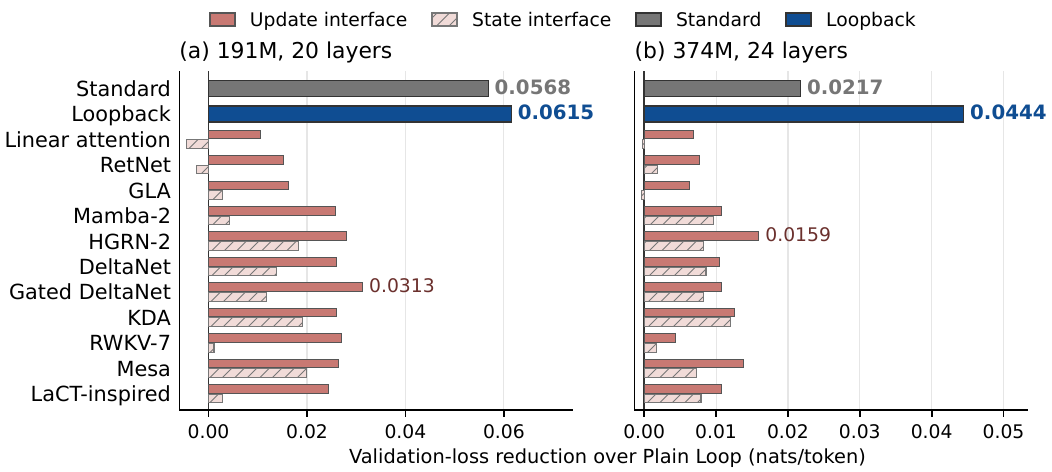}
\caption{\textbf{Depth-memory readers.} ClimbMix validation-loss reduction over Plain Loop at
(a) 191M with 20 layers and (b) 374M with 24 layers for the update interface (solid) and the state interface (hatched) of
each reader; the standard Transformer and Loopback are references.}
\label{fig:memory-readers}
\end{figure}

\Needspace{10\baselineskip}
The update interface gives lower validation loss in all 22 family--scale
comparisons (\cref{fig:memory-readers,tab:readers-compressed}). Every
update-based reader improves on Plain Loop, but the best readers still trail
Loopback by 0.0302 nats at 191M and 0.0285 at 374M. These results favor
updates over states as the memory value. Fixed-state recurrent cores keep
one practical advantage for long loops: each new write is absorbed without
rereading the history, and the state does not grow with the visit count. We
have not tested whether optimized recurrent kernels turn this into lower memory
traffic or higher throughput in deeply looped models.

\Needspace{4\baselineskip}
\subsection{Efficiency: compute, throughput, and memory}
\label{sec:efficiency}

Loopback's residual aggregation adds only 0.04--0.08\% forward FLOPs over Standard at matched executed depth
in the three measured configurations (\cref{fig:efficiency,tab:compute}).
These extra operations normalize and
score stored updates, form their weighted sum, and
apply decay and mixture replacement; they introduce no Q/K/V projections or
extra block evaluations. With our fused implementation, 3B training
throughput is 6\% below Standard's
on the same eight H100 GPUs with the same batch layout. Detailed runtime and
memory measurements are reported in \cref{app:compute}.

% Figure: compute and efficiency (Section 6.5). Generated by figure/efficiency/fig_efficiency.py from the measured
% training logs (train_steptimes.json, tim_fused.json) and the FLOPs count (flops_per_token.json); tab:compute holds
% the numbers for every scale. AttnRes (blockstate, 10-21) comes from its own 20-TPP log; FLOPs from flops_blockstate.py. Regenerate fig_efficiency.pdf after any change to those inputs.
\begin{figure}[!htb]
\setlength{\abovecaptionskip}{5pt}
\centering
\includegraphics[width=\linewidth]{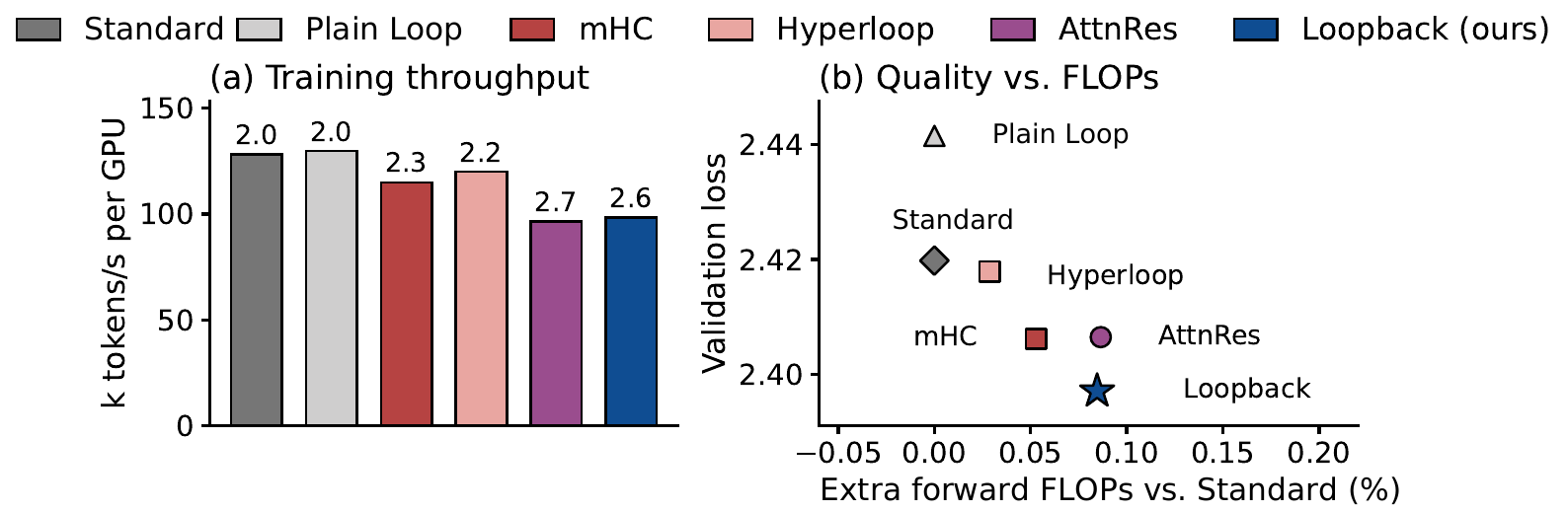}
\caption{\textbf{Compute and efficiency at 374M, 24 layers.} Loopback and AttnRes run fused read kernels.
(a) Per-GPU training throughput on eight H100s; bar labels give hours for the 20-TPP budget. (b) Validation
loss versus forward FLOPs per token in excess of the standard baseline's.}
\label{fig:efficiency}
\end{figure}
\endgroup

\FloatBarrier
\section{Related work}
\label{sec:related}

\textbf{Parameter sharing and residual reuse.}
Models with shared depth use weight tying
\citep{dehghani2018universal,lan2019albert}, relaxed tying with low-rank adaptations
\citep{bae2025relaxed}, latent reasoning \citep{saunshi2025reasoning,geiping2026scaling},
and adaptive recursion \citep{bae2026mixture}. Hyperloop learns connections between
loop visits \citep{zeitoun2026hyperloop}. Residual pathways use gating, addition,
and concatenation
\citep{srivastava2015highwaynetworks,he2016deep,huang2018denselyconnectedconvolutionalnetworks},
scaling \citep{bachlechner2021rezero,wang2024deepnet}, depth mixtures
\citep{bapna2018training}, multi-stream mixing \citep{zhu2025hyper,xie2025mhc}, and
attention-based aggregation \citep{team2026attention,luo2026delta,luo2026multi}.
Storing updates is not new; what we add is the split of update coefficients
into input strength, total weight, and allocation, and an account of how each
design constrains them.

\textbf{Attention and recurrent memory.}
Sequence memory includes attention and additive fast weights
\citep{vaswani2017attention,katharopoulos2020transformers,schlag2021linear},
retention and gating \citep{sun2023retentive,yang2023gated,qin2024hgrn2}, structured
state-space dynamics \citep{gu2021efficiently,gu2023mamba,dao2024transformers}, delta-rule
corrections
\citep{yang2024parallelizing,yang2025gated,team2025kimi,peng2025rwkv}, and
optimization-based memory \citep{von2023uncovering,zhang2026test}.
We adapt these operations along depth, retaining each block's token attention;
results concern residual interfaces, not the original architectures' efficiency.

\Needspace{7\baselineskip}
\section{Conclusion}

We present Loopback, a residual design for parameter-efficient looped Transformers. We connect residual computation with linear attention and recurrent memory by treating successive block applications as memory steps along depth: each application writes new computation, which later blocks access through the residual connection. Loopback stores each block's update separately and reads this history with attention. The read starts from geometrically decayed weights and redistributes them according to content without changing their sum; the weighted updates are then added to the original loop input. Across the 191M, 374M, 778M, 1.4B, and 3B scales, Loopback improves aggregate quality over matched-depth standard Transformer baselines while using 40--75\% of their Transformer parameters, excluding token embeddings and the output head. In looped models, how the results of shared computation are retained and reused matters as much as which blocks are shared.

\section{Limitations and future work}
\label{sec:limitations}
\label{app:limitations}

Our evidence is limited to ClimbMix pretraining at the tested scales and
budgets, with single-run large-scale comparisons. The input coefficient is
fixed at one; we have not tried learned or layer-specific input injection.
Explicit history adds memory traffic and activations, so parameter savings
need not reduce latency or peak memory. Extra visits have not improved
validation loss. Next steps are other pretraining corpora, controlled
comparisons of the reader interfaces, compact memories, and training with a
variable visit count.

\section*{Ethics statement}
This work studies the architecture of language models. All models have at most
3B parameters and are trained on ClimbMix, a public pretraining corpus, with
evaluation on public benchmarks; no new data were collected and no human
subjects were involved. The models are research artifacts that inherit the
biases of their training corpus and are not intended for deployment. The method
reduces stored parameters at matched executed depth and does not alter the risks
associated with language models in general.

\Needspace{8\baselineskip}
\bibliographystyle{iclr2027_conference}
\begingroup
\raggedright
\urlstyle{same}
\setlength{\bibsep}{2pt plus 1pt}
\bibliography{references}
\endgroup
\clearpage
\appendix
\section{Proofs for the depth-memory formulation}
\label{app:framework-details}

\subsection{Coefficient ordering for cumulative states}
\label{app:cumulative-states}

For ordinary additive states
$\bm h_j^{\mathrm{add}}=\bm h_0+\sum_{i<j}\vd_i$, a nonnegative normalized
read gives
\begin{equation}
    \sum_{j=0}^{s+1}\rho_j\bm h_j^{\mathrm{add}}
       =\bm h_0+\sum_{i=0}^{s}\left(\sum_{j=i+1}^{s+1}\rho_j\right)\vd_i,
       \qquad \rho_j\ge0,\quad\sum_j\rho_j=1.
    \label{eq:state-to-delta}
\end{equation}
The equality follows by exchanging the finite sums: update $\vd_i$ appears
in every state with $j>i$. Hence
$c_{s,i}=\sum_{j=i+1}^{s+1}\rho_j$ and
$1\ge c_{s,0}\ge\cdots\ge c_{s,s}\ge0$. Reading block updates separately removes
this cumulative-sum restriction, but does not make the updates independent.

For a reweighted trajectory $\bm v_j=a_j\bm h_0+\sum_i C_{j,i}\vd_i$,
the induced update weights are $\bm\rho^\top\bm C$; the ordering above
need not hold.

\subsection{Proof of the coefficient expansion}

Repeated substitution in \cref{eq:delta-transport} gives
\begin{equation}
    \bm M_{s+1}=\bm M_0\bm A_0\cdots\bm A_s
                   +\sum_{i=0}^{s}\vd_i\bm b_i^\top
                       \bm A_{i+1}\cdots\bm A_s.
    \label{eq:operator-unroll-proof}
\end{equation}
The base case $s=0$ is the update rule. Multiplying the previous expansion
by $\bm A_s$ and adding $\vd_s\bm b_s^\top$ proves the induction step.
The factors retain their order, and empty products are identity matrices. Substituting
$\bm M_0=\bm h_0\bm a_0^\top$ and multiplying by $\bm r_s$ yields
\cref{eq:history-coefficients}. A separate position write
$\bm e_i\bm t_i^\top$ contributes
$(\bm t_i^\top\bm A_{i+1}\cdots\bm A_s\bm r_s)\bm e_i$
by the same calculation.

The identity is evaluated on the actual trajectory. Keys, queries, and gates
may depend on content; the resulting coefficients are still scalar. Projections
of the values themselves require operator-valued coefficients; arbitrary
nonlinear value maps are outside this expansion.
\section{Residual adapters and reader implementations}
\label{app:operator-mapping}

\subsection{Stream transport and explicit history}

The block-level interface distinguishes a block update $\vd_s$ from a full
block output $f_k(\bm x_s)$, where $k=s\bmod K$. Our iHC adapter writes the block update.
Hyper-Connections, mHC, and Hyperloop instead operate at visit boundaries:
Hyper-Connections writes the whole-visit update, while mHC and Hyperloop
write the full stack output together with a position value. Applying
\cref{eq:delta-transport} to the stack map $F$ gives
$\bm A_r=\bm W_r+\bm p_r\bm b_r^\top$ for the latter two, with the
additional write $\bm e_r\bm b_r^\top$; $r$ indexes visits here.
In value-by-stream storage, mHC uses
$\bm W_r=\bm H_{\mathrm{res}}^\top$ and constrains this transport by Sinkhorn
normalization; the additional input-write path is not part of that constraint.
Hyperloop uses diagonal transport. iHC instead fixes
$\bm A_s=\bm W_s=\vI$ and learns nonnegative write gates, input reads,
and the final-coda read. Its original rule surrounds the attention and MLP
computations separately; our adapter surrounds each complete block and
writes $\vd_s=f_k(\bm x_s)-\bm x_s$.

For a history bank, appending $\vd_s$ uses $\bm A_s=[\vI,\bm0]$ and a write
vector selecting the new storage location. Appending the full block output instead uses
$\bm A_s=[\vI,\bm p_s]$. The sources and placements used by AttnRes, MHAR,
and Loopback are specified in \cref{app:190m-evaluation}.

\subsection{Normalized attention and recurrent memory}
\label{app:linear-memory}

\paragraph{Finite-feature attention.}
Linear attention summarizes values $\vv_i=\vv_i^{(\chi)}$ in sufficient
statistics \citep{katharopoulos2020transformers,schlag2021linear}.
For a finite feature map $\phi:\R^{d_k}\to\R^m$ with
$\kappa(\vq,\vk)=\phi(\vq)^\top\phi(\vk)\ge0$, associativity gives
\begin{equation}
    \bm o_s=\frac{\sum_{i=0}^{s}\kappa(\vq_s,\vk_i)\vv_i}
                       {\sum_{i=0}^{s}\kappa(\vq_s,\vk_i)}
       =\frac{\vS_s\phi(\vq_s)}{\bm z_s^\top\phi(\vq_s)},\qquad
    \vS_s=\sum_{i=0}^{s}\vv_i\phi(\vk_i)^\top,\quad
    \bm z_s=\sum_{i=0}^{s}\phi(\vk_i).
    \label{eq:linear-depth-read}
\end{equation}
Updating $\vS_s\in\R^{d\times m}$ and $\bm z_s\in\R^m$ avoids storing
past values for either $\chi$. The denominator must be positive; this
finite-feature identity is not a general softmax equivalence.

A destination-dependent softmax query changes earlier scores, so its previous
read alone generally cannot summarize the history. Our evaluated recurrent
adapters instead use the unnormalized memory update in \cref{eq:memory}, omitting the
kernel normalizer and allowing signed read coefficients.

\paragraph{Decay and key-specific correction.}
\label{app:memory-corrections}

The delta rule corrects a memory prediction; it does not denote the block
output-minus-input transformation. At block granularity, its target can be
either a full block output or a block update. For these families, let
$\bar{\vS}_i=\vS_{i-1}\bm D_i$:
\begin{equation}
    \vS_i=\bar{\vS}_i+
       \eta_i(\vv_i-\bar{\vS}_i\vk_i)\vk_i^\top,\qquad
    \bm D_i=
    \begin{cases}
        \vI & \text{DeltaNet},\\
        \gamma_i\vI & \text{Gated DeltaNet},\\
        \Diag(\bm g_i) & \text{KDA}.
    \end{cases}
    \label{eq:decay-correction}
\end{equation}
Decay acts before correction toward the new value
\citep{yang2024parallelizing,yang2025gated,team2025kimi}.
Thus KDA uses
$\bm W_i=\Diag(\bm g_i)(\vI-\eta_i\vk_i\vk_i^\top)$ in value-by-key
orientation; reversing the factors generally changes the operator.
RWKV-7 subtracts a rank-one correction from diagonal transport. Its normalized
removal direction $\bm\kappa_s$ and coordinatewise strength $\bm a_s$ need
not align with the write key $\vk_s$ \citep{peng2025rwkv}.

\subsection{Regression readers and their additional state}
\label{app:regression}

The regression readers in \cref{tab:families} require additional state:
Mesa tracks inverse covariance, while the LaCT-inspired reader
revisits a stored prefix.

\paragraph{Mesa.}
Let $\bm K=[\vk_0,\ldots,\vk_s]$, $\bm V=[\vv_0,\ldots,\vv_s]$,
$\bm\Omega=\Diag(\omega_0,\ldots,\omega_s)$ with $\omega_i\ge0$, and $\mu>0$.
Differentiating the weighted ridge objective gives
\begin{align}
    \bm S_s
      &=\arg\min_{\bm W}\frac12
         \|(\bm W\bm K-\bm V)\bm\Omega^{1/2}\|_F^2
                         +\frac\mu2\|\bm W\|_F^2\notag\\
      &=\bm V\bm\Omega\bm K^\top
                    (\bm K\bm\Omega\bm K^\top+\mu\vI)^{-1}.
    \label{eq:ridge}
\end{align}
Its recursive form starts from $\vS_{-1}=0$ and $\bm R_{-1}=\mu^{-1}\vI$:
\begin{align}
    \bm\ell_i&=\frac{\omega_i\bm R_{i-1}\vk_i}
                    {1+\omega_i\vk_i^\top\bm R_{i-1}\vk_i},\notag\\
    \vS_i&=\vS_{i-1}(\vI-\vk_i\bm\ell_i^\top)+\vv_i\bm\ell_i^\top,
    \qquad
    \bm R_i=\bm R_{i-1}-\bm R_{i-1}\vk_i\bm\ell_i^\top.
    \label{eq:mesa-operator-map}
\end{align}
The rank-one inverse identity yields
$\bm R_i=(\mu\vI+\sum_{j\le i}\omega_j\vk_j\vk_j^\top)^{-1}$.
Substituting this into the ridge normal equations gives the update for $\vS_i$.
Hence $\bm W_i=\vI-\vk_i\bm\ell_i^\top$ and $\bm b_i=\bm\ell_i$,
with $\bm R_i$ as control state. Our implementation evaluates the equivalent
weighted Gram solve.

\paragraph{LaCT-inspired reader.}
Let $\bm K_i=[\vk_0,\ldots,\vk_i]$, $\bm V_i=[\vv_0,\ldots,\vv_i]$,
and $\bm E_i=\Diag(\eta_0,\ldots,\eta_i)$. A gradient step on the
weighted prefix prediction error, followed by the implemented normalization, is
\begin{align}
    \overline{\vS}_i
       &=\vS_{i-1}(\vI-\bm K_i\bm E_i\bm K_i^\top)
                             +\bm V_i\bm E_i\bm K_i^\top,\notag\\
    \vS_i&=\nu_i\overline{\vS}_i,\qquad
       \nu_i=(\|\overline{\vS}_i\|_F^2+\varepsilon^2)^{-1/2},
       \qquad \vS_{-1}=0.
    \label{eq:lact-prefix-map}
\end{align}
Because old values are written again, $\vS_i$ alone is insufficient for a
single-new-value update. Set $\bm M^-=[\vS_{i-1},\bm V_{<i}]$ and
$\bm G_i=\vI-\bm K_i\bm E_i\bm K_i^\top$. The augmented operators
\begin{equation}
    \bm W_i=
       \begin{bmatrix}
          \nu_i\bm G_i&0&0\\
          \nu_i\bm E_{<i}\bm K_{<i}^\top&\vI&0
       \end{bmatrix},
    \qquad
    \bm b_i^\top=[\nu_i\eta_i\vk_i^\top,\bm0^\top,1]
    \label{eq:lact-augmented-map}
\end{equation}
give $\bm M^+=\bm M^-\bm W_i+\vv_i\bm b_i^\top=[\vS_i,\bm V_i]$
by direct multiplication, with read $[\vq_i;\bm0]$ selecting $\vS_i\vq_i$.
The initial augmented state is $[0,\varnothing]$; prefix blocks are empty at $i=0$.
This adapter uses linear predictions over the cumulative prefix, not the
published LaCT layer. Its Frobenius normalization makes the update nonlinear,
preserves the value span, and remains in the derivative graph.

\subsection{Visit-level reader interfaces}
\label{app:readers}

The two interfaces in \cref{tab:readers-compressed} read at visit boundaries
and initialize to Plain Loop. Both use unprojected values, L2 normalization
with an epsilon floor, and no learned Q/K/V projections.

\subsubsection{Cumulative states and whole-visit updates}
\label{app:algebra}
\label{app:visit-granularity}

Write $\widetilde{\bm h}_{(r+1)K}$ for the output of visit $r$ before its
boundary combination. The paired visit update is
\begin{equation}
    \bm\Delta_r=\widetilde{\bm h}_{(r+1)K}-\bm h_{rK}
              =\sum_{k=0}^{K-1}\vd_{rK+k}.
    \label{eq:visit-granularity}
\end{equation}
The equality telescopes under within-visit addition, as used here;
block-level reweighting would also enter the visit difference.

The state interface stores $\bm h_0$ and each visit output before its boundary
blend. Denote these stored states by $\bm v_j$. Within each reader head,
\[
    \vk_j=\operatorname{L2Norm}(\bm v_j),\qquad
    \vv_j=\bm v_j,\qquad
    \vq=\operatorname{L2Norm}(\widetilde{\bm h}_{(r+1)K}).
\]
The current key and query coincide, so their
inner product is one when the normalization floor is inactive. The memory
output $\bm o_r$ is combined with the current proposal as
\begin{equation}
    \bm h_{(r+1)K}=(1-\zeta)\widetilde{\bm h}_{(r+1)K}
                                      +\zeta\bm o_r,
    \qquad \zeta_{\mathrm{init}}=0.
    \label{eq:state-blend}
\end{equation}

The update interface stores $\bm\Delta_0,\ldots,\bm\Delta_r$, with
\[
    \vk_j=\operatorname{L2Norm}(\bm\Delta_j),\qquad
    \vv_j=\bm\Delta_j,\qquad
    \vq=\operatorname{L2Norm}(\widetilde{\bm h}_{(r+1)K}).
\]
Its output is
\begin{equation}
    \bm h_{(r+1)K}=(1-\zeta)\widetilde{\bm h}_{(r+1)K}
                          +\zeta(\bm h_0+\bm o_r),
    \qquad \zeta_{\mathrm{init}}=0.
    \label{eq:visit-blend}
\end{equation}
In both interfaces $\zeta$ is an unconstrained learned scalar initialized to
zero, recovering Plain Loop.

\subsubsection{Gates and memory evaluation}

In our adapters, gates are content-independent learned parameters indexed by
junction, head, and, when required, channel. Key-addressed families write along $\vk_j$; HGRN-2 writes
the gate complement instead. For RWKV-7, the removal direction
$\bm\kappa_j$ in \cref{tab:families} is the normalized visit update in the
update interface and the normalized difference of adjacent stored states in
the state interface. Mesa and the LaCT-inspired reader follow
\cref{app:regression}, including the latter's prefix normalization.

The implementations evaluate short histories using the factorization
$\vS_s=\sum_i\vv_i\bm w_{s,i}^\top$, where $\bm w_{s,i}$ is a write vector
after subsequent transitions, or replay the required prefixes. These evaluate
the stated reads without establishing an end-to-end recurrent speedup.

\section{Loopback implementation and derivations}
\label{app:implementation}

\subsection{Mixture replacement and initialization}

After appending the newest block update and applying anchored decay at visit
ends, write the current proposal as
$\bm h^{\mathrm{base}}_{s+1}=\bm h_0+\sum_i\bar c_{s,i}\vd_i$.
The implementation reaches the target in \cref{eq:instance} by adding only
the coefficient difference:
\begin{equation}
    \bm h_{s+1}=\bm h^{\mathrm{base}}_{s+1}
                       +\sum_i(c_{s,i}-\bar c_{s,i})\vd_i.
    \label{eq:replacement}
\end{equation}
The implementation uses skip parameter $\beta=1-\lambda$ and tracks the
actual coefficients after prior reads to avoid double-counting contributions.
The explicit input path has unit gain, but the full Jacobian also includes
the input dependence of the updates and routing weights.

To make the initial correction vanish, target coefficients are evaluated as
\begin{equation}
    \bm c_s=\bm c_s^0+B_s\left[
       \operatorname{softmax}(\bm z_s+\log\bm c_s^0)
         -\operatorname{softmax}(\log\bm c_s^0)\right].
    \label{eq:centered}
\end{equation}
For a positive unfloored prior this equals
$B_s\operatorname{softmax}(\bm z_s+\log\bm c_s^0)$ in exact arithmetic.
Zero queries cancel the bracket; causal induction then recovers the decay-only
trajectory of \cref{sec:outer-decay}. The first read has only one source and
requires no query.
The $(KR-1)d$ learned parameters total 11,520 at $K=4$, $R=4$, $d=768$ (the 191M setting).

The actual log prior uses $\log\max(c^0_{s,i},10^{-8})$ in both softmax
terms. If the floor activates, the centered form still preserves the total
weight in exact arithmetic and cancels at zero queries, but its coefficients
need not remain nonnegative. All reported configurations satisfy
$\lambda^R>10^{-8}$, so the floor is inactive.

\subsection{Allocation gradient}
\label{app:allocation-gradient}

With the update bank, prior, and total $B$ fixed, and the prior floor inactive
so that $\bm c=B\operatorname{softmax}(\bm z+\log\bm c^0)$ exactly, write $\pi_i=c_i/B$ and
$\bm\mu=\sum_j\pi_j\vd_j$, so $\bm h=\bm h_0+B\bm\mu$.
The softmax derivative
$\partial\pi_j/\partial z_i=\pi_j(\mathbf1_{i=j}-\pi_i)$ gives
$\partial\bm h/\partial z_i=c_i(\vd_i-\bm\mu)$.
Contraction with $\bm g=\nabla_{\bm h}\mathcal L$ yields
\begin{equation}
    \frac{\partial\mathcal L}{\partial z_i}
       =c_i\bm g^\top(\vd_i-\bm\mu),\qquad
    \sum_i\frac{\partial\mathcal L}{\partial z_i}=0 .
    \label{eq:allocation-gradient}
\end{equation}
The sum vanishes because $\sum_i c_i\vd_i=B\bm\mu$.
This local gradient redistributes a fixed total according to each update's
alignment with the downstream loss gradient. Actual score changes are
coupled through the query; full training also changes the keys and updates.

\FloatBarrier
\section{Training and evaluation protocols}
\label{app:190m}

\Cref{tab:scale-configs} lists the architecture, batch composition, and budget
at every scale. We give the shared recipe and evaluation protocol once,
then summarize larger-scale differences. The per-GPU sequence counts in \cref{tab:compute}
are these training micro-batches, and timing runs use the same micro-batch and
accumulation.

% Written from the run recipes (sigma_runs_climb4k/*_recipe.txt; 1.4B: the N4-3 t1_ runs of 2026-10-03) and the model
% configs (3B: the N4-3 b3n/b3p runs of 2026-10-15/25, configs_d3072x24). SwiGLU widths are confirmed by the Transformer parameter counts of Tables 2 and 15
% (per block: 4d^2 attention + 3 d w feed-forward + 2d norms; e.g. 374M: 24 x 12,847,104 = 308.33M).
\begin{table}[H]
\caption{\textbf{Configuration at each scale.} Blocks and heads apply to Standard
and to every looped arm; head width is 64 throughout. Sequences per optimizer
step are the per-GPU micro-batch times gradient-accumulation steps, on 8
ranks, at 4,096 tokens per sequence. Peak learning rates are Standard / looped
arms.}
\label{tab:scale-configs}
\centering\scriptsize
\setlength{\tabcolsep}{2.5pt}
\renewcommand{\arraystretch}{1.06}
\begin{tabular*}{\linewidth}{@{\extracolsep{\fill}}lrrrrlrrrrl@{}}
\toprule
Scale & Width & Blocks & Heads & SwiGLU & Loopback & Seq./step & Micro$\times$acc. & Steps & Tokens & LR ($\times10^{-3}$) \\
\midrule
191M & 768 & 20 & 12 & 2,048 & $\looptopology{2}{4}{4}{2}$ & 128 & $16\times1$ & 7,280 & 3.82B & 1 / 3 \\
374M & 1,024 & 24 & 16 & 2,816 & $\looptopology{2}{10}{2}{2}$ & 256 & $8\times4$ & 7,120 & 7.47B & 2 / 2 \\
778M & 1,536 & 24 & 24 & 4,096 & $\looptopology{2}{8}{2}{6}$ & 512 & $8\times8$ & 7,440 & 15.60B & 1 / 1.5 \\
1.4B & 2,048 & 25 & 32 & 5,632 & $\looptopology{3}{8}{2}{6}$ & 960 & $8\times15$ & 7,200 & 28.31B & 1 / 1.5 \\
3B & 3,072 & 24 & 48 & 8,192 & $\looptopology{6}{6}{2}{6}$ & 1,920 & $4\times60$ & 7,600 & 59.77B & 1 / 1 \\
\bottomrule
\end{tabular*}
\end{table}

\subsection{Architecture and training protocol}
\label{app:190m-settings}
\label{app:larger-settings}
\label{app:1024-settings}
\label{app:1536-settings}
\label{app:2048-settings}
\label{app:2900-settings}

Scale names use the standard baseline's total parameters; reported Transformer
counts exclude embeddings and the vocabulary head. We use the 20-TPP training
setting, with the same token budget for all models within each comparison.
At 191M, all nine models train from scratch for 7,280 steps using
\cref{tab:190m-architecture,tab:190m-training}, differing only in topology,
residual rule, and peak learning rate. An additional eight-block standard Transformer baseline
matches Plain Loop's stored parameters (56.64M), not its twenty applications;
under the same protocol and tokens, its loss is 2.7376 versus Plain Loop's 2.6802.
Budgets are rounded to a 20-step grid. During the final tenth of training,
the learning-rate multiplier is $1-\sqrt{u}$ for decay progress $u\in[0,1]$.
All results use the final checkpoint.

The 374M, 778M, 1.4B, and 3B runs use the same optimizer, tokenization,
normalization, schedule, and evaluation protocols.
\Cref{tab:scale-configs} gives their widths, topologies, batches, budgets, and
peak learning rates. The 1.4B and 3B runs use training shards 0--559; smaller
scales use shards 0--239. Within each comparison, training tokens and executed
depth are matched. The depth-sweep variants are specified in \cref{app:depth}.

\begin{table}[H]
\caption{\textbf{191M architecture.}
Looped topology: $\looptopology{P}{K}{R}{C}$.}
\label{tab:190m-architecture}
\centering\small
\setlength{\tabcolsep}{5pt}
\renewcommand{\arraystretch}{1.12}
\begin{tabularx}{\linewidth}{@{}l>{\raggedright\arraybackslash}X@{}}
\toprule
Setting & Value \\
\midrule
Standard structure & Twenty independent blocks \\
Looped structure & $\looptopology{2}{4}{4}{2}$; eight distinct blocks, twenty applications \\
Attention & Full-sequence causal attention; 12 heads of width 64; no GQA \\
Hidden / feed-forward width & 768 / 2,048; SwiGLU feed-forward layers \\
Normalization & Pre-RMSNorm; $\epsilon=10^{-6}$ \\
Position encoding & RoPE; $\theta=10^6$ \\
Context / vocabulary & 4,096 tokens / 32,000 Llama tokens \\
Embedding / vocabulary head & Untied; omitted from reported Transformer counts \\
Trunk initialization & Linear and embedding weights: $\mathcal N(0,0.02^2)$ \\
Other trunk options & Bias-free attention projections; no QK normalization,
logit softcap, or zero-initialized residual-output projections \\
\bottomrule
\end{tabularx}
\end{table}

\begin{table}[H]
\caption{\textbf{Training recipe at the 20-TPP endpoint.}
Only peak learning rates differ across arms.}
\label{tab:190m-training}
\centering\small
\setlength{\tabcolsep}{5pt}
\renewcommand{\arraystretch}{1.12}
\begin{tabularx}{\linewidth}{@{}l>{\raggedright\arraybackslash}X@{}}
\toprule
Setting & Value \\
\midrule
Training data & ClimbMix \citep{diao2026nemotron}; training shards 0--239 \\
Input construction & Online tokenization; concatenated documents packed into
fixed 4,096-token sequences \\
Global batch & 128 sequences; $8$ GPU ranks $\times16$ sequences per rank \\
Gradient accumulation & 1; 524,288 input tokens per optimizer step \\
Training budget & 7,280 steps; 3,816,816,640 input tokens \\
Optimizer & AdamW; $(\beta_1,\beta_2)=(0.9,0.95)$; $\epsilon=10^{-15}$ \\
Weight decay / gradient clip & 0 / global norm 1.0 \\
Peak learning rate & Standard: $10^{-3}$; all looped arms: $3\times10^{-3}$ \\
Learning-rate schedule & Linear warmup: 128 steps; stable peak: 6,424 steps;
square-root decay to zero: 728 steps \\
Hardware allocation & One node with eight GPU ranks for every arm \\
Mixed precision & BF16 parameter computation; FP32 gradient reductions \\
Execution & FSDP; no activation checkpointing or model compilation \\
\bottomrule
\end{tabularx}
\end{table}

\Needspace{5\baselineskip}
\subsection{Residual rules and evaluation}
\label{app:190m-evaluation}

Trunk weights are shared across visits; connections and queries are untied
where required. The mHC, Hyper-Connections, and Hyperloop adapters act at
visit boundaries with four streams; mHC uses 20 Sinkhorn iterations and
Hyperloop uses zero residual bias. iHC wraps each middle block with four
streams, reading the block input through gates before it and writing the block
update back through gates after it; it uses magnitude 2, gate offset
$10^{-6}$, and a learned final readout.
MHAR reads loop states at visit boundaries with eight heads and an outer
blend. AttnRes reads the loop input and block states before token attention
and before the MLP in each loop block, and once at loop exit, using Block
AttnRes with $N=8$. Loopback reads block
updates after each middle block with one head and $\lambda=1/4$.
The queries and gates of MHAR, AttnRes, and Loopback start at zero.

\paragraph{Held-out language modeling.}
Validation uses disjoint ClimbMix shard 3583. As in training, documents are
tokenized with beginning-of-sequence markers, concatenated, and packed into
4,096-token sequences. All models use the first 1,024 sequences and their final
checkpoints, evaluated in BF16. Loss averages next-token cross-entropy over 4,095 predicted
positions per sequence after shifting labels. It is reported in nats;
perplexity is its exponential.

\Needspace{10\baselineskip}
\paragraph{Downstream tasks.}
The same final checkpoints are evaluated using lm-evaluation-harness 0.4.12
\citep{gao2021framework}, zero-shot prompts, full evaluation sets, and 1,000
bootstrap iterations; every prompt starts with the beginning-of-sequence token,
as training prepends it to every document. HellaSwag, OpenBookQA, ARC-Easy, and
ARC-Challenge use length-normalized accuracy; the other accuracy tasks use the
harness's standard accuracy. MMLU is the harness aggregate over 57 subjects and
counts as one task in the unweighted thirteen-task mean. WikiText word
perplexity is reported separately and excluded from this mean.

\FloatBarrier
\Needspace{12\baselineskip}
\subsection{Learning-rate selection}
\label{app:learning-rates}

\noindent\begin{minipage}{\linewidth}
At 191M, Standard's lowest loss among the four peak rates tested at 20~TPP
is attained at $10^{-3}$ (\cref{tab:dense-lr}); looped models use their best
tested rate, $3\times10^{-3}$.
% Additional 190M tables. Numeric rows come from LoopResidual/results/paper_tables_190m/mk_tables.py; captions are manuscript prose.

\captionof{table}{\textbf{191M Standard learning-rate comparison.}
ClimbMix validation loss after 20~TPP on the same machine and training recipe.
Bold marks the lowest loss.}
\label{tab:dense-lr}
\centering\footnotesize
\begin{tabular}{@{}lrrrr@{}}
\toprule
Peak learning rate & $5\times10^{-4}$ & $7\times10^{-4}$ & $10^{-3}$ & $2\times10^{-3}$ \\
\midrule
ClimbMix loss & 2.6666 & 2.6420 & \textbf{2.6234} & 2.6245 \\
\bottomrule
\end{tabular}
\end{minipage}\par\addvspace{8pt}

For the final 20-TPP runs, 374M uses $2\times10^{-3}$ for all arms, selected
for Standard from $\{1,1.5,2\}\times10^{-3}$. At 778M, Standard inherits
$10^{-3}$; Loopback uses $1.5\times10^{-3}$ after a check against
$2\times10^{-3}$. At 1.4B, Standard uses $10^{-3}$ from
$\{0.5,1,1.5,2\}\times10^{-3}$, and Loopback uses $1.5\times10^{-3}$
from $\{1,1.5\}\times10^{-3}$. At 3B, both use $10^{-3}$, from
$\{0.6,1,1.5\}\times10^{-3}$ for Standard and
$\{0.7,1,1.5\}\times10^{-3}$ for Loopback.
Screening used 5~TPP at 374M/778M and 2~TPP at 1.4B/3B; 1.4B additionally
compared $10^{-3}$ and $1.5\times10^{-3}$ at 20~TPP.
Loopback's 1.4B and 3B rate checks used $\looptopology{3}{10}{2}{2}$ and
$\looptopology{2}{10}{2}{2}$, respectively; the final 3B topology was chosen
in a separate 5-TPP screen. Short runs apply their own final decay, and
continuations resume from the pre-decay checkpoint.
\FloatBarrier
\Needspace{32\baselineskip}
\subsection{Full general-benchmark results at 1.4B and 3B}
\label{app:large-general}

\Cref{tab:scale-large} gives the taskwise scores underlying the large-scale
comparison, using the same beginning-of-sequence prompts and metric conventions
as the smaller-scale evaluation.

% Generated by figures/make_scale_large_table.py; edit the CSV/JSON inputs, then regenerate.
% Source versions and BOS evaluation protocol are recorded in figures/scale_large_metadata.json.
\begin{table}[H]
\setlength{\abovecaptionskip}{4pt}
\caption{\textbf{Standard and Loopback at 1.4B and 3B.} Results after 20~TPP, one run per arm.
Task scores are zero-shot accuracies (\%) with beginning-of-sequence prompts; bold marks the better
result within each scale. Parameter shares compare Transformer parameters, excluding embeddings and
the vocabulary head.}
\label{tab:scale-large}
\centering\small
\setlength{\tabcolsep}{4pt}
\renewcommand{\arraystretch}{1.04}
\begin{tabular*}{\linewidth}{@{\extracolsep{\fill}}lr>{\columncolor{paperTeal!7}}rr>{\columncolor{paperTeal!7}}r@{}}
\toprule
& \multicolumn{2}{c}{1.4B, 25 layers, 28.31B tokens} & \multicolumn{2}{c@{}}{3B, 24 layers, 59.77B tokens} \\
\cmidrule(lr){2-3}\cmidrule(l){4-5}
& Standard & \textbf{Loopback} & Standard & \textbf{Loopback} \\
\midrule
Topology & -- & $\looptopology{3}{8}{2}{6}$ & -- & $\looptopology{6}{6}{2}{6}$ \\
Transformer params (M) & 1,284.61 & 873.57 & 2,718.06 & 2,038.58 \\
Relative to Standard & 100.0\% & 68.0\% & 100.0\% & 75.0\% \\
\midrule
ClimbMix loss $\downarrow$ & 2.1283 & \textbf{2.1073} & 2.0005 & \textbf{1.9849} \\
\midrule
PIQA & 73.88 & \textbf{74.48} & \textbf{77.04} & 75.90 \\
HellaSwag & 55.02 & \textbf{56.43} & 62.84 & \textbf{63.72} \\
WinoGrande & 56.51 & \textbf{56.99} & 57.22 & \textbf{58.48} \\
OpenBookQA & \textbf{37.80} & 36.60 & 40.40 & \textbf{41.80} \\
CommonsenseQA & \textbf{23.34} & 18.67 & 19.90 & \textbf{21.87} \\
BoolQ & 62.51 & \textbf{63.24} & 62.23 & \textbf{64.95} \\
RACE & \textbf{34.64} & 33.49 & 35.12 & \textbf{36.17} \\
ARC-Easy & 66.16 & \textbf{67.00} & 69.19 & \textbf{71.13} \\
ARC-Challenge & 36.35 & \textbf{37.03} & 40.53 & \textbf{40.96} \\
MMLU (57 subjects) & 24.70 & \textbf{26.81} & 26.26 & \textbf{28.86} \\
ANLI-R1 & 31.30 & \textbf{34.80} & 31.00 & \textbf{32.90} \\
RTE & \textbf{57.40} & 55.23 & \textbf{61.01} & 56.68 \\
LAMBADA & 42.64 & \textbf{43.72} & 48.50 & \textbf{49.45} \\
\midrule
Thirteen-task mean (\%) $\uparrow$ & 46.33 & \textbf{46.50} & 48.56 & \textbf{49.45} \\
WikiText perplexity $\downarrow$ & 21.09 & \textbf{20.29} & 17.28 & \textbf{16.74} \\
\bottomrule
\end{tabular*}
\par\vspace{2pt}{\footnotesize At 3B, the bootstrap standard error of each model's thirteen-task mean is 0.40 percentage points.}
\end{table}
\FloatBarrier

\FloatBarrier
\section{Run-to-run variation and memory-reader results}
\subsection{Run-to-run variation}
\label{app:seeds}

\Cref{tab:seeds} reports two training seeds for every 191M arm except AttnRes
and four arms at 374M (24 layers). Each seed sets both the initialization
and the data order. At 191M, Loopback has the
lowest validation loss among the tested looped models in both runs. A third
seed for the standard Transformer, Plain Loop, mHC, and Loopback keeps this ordering
(Loopback $-0.0110$ against mHC and $-0.0716$ against Plain Loop). Its paired loss
differences from the standard Transformer are $-0.0047$, $+0.0011$, and
$-0.0064$ (mean $-0.0033$), so that comparison is not consistent across seeds.
At 374M, Loopback has the lowest validation loss among the four tested models
in both seeds.

% Numeric results originated in figure/depth_sweep/make_seed_table.py from seed_runs.json.
% The 374M rows are restored from manuscript commit c96bcb7; formatting is maintained here.
\begin{table}[H]
\caption{\textbf{Validation-loss variation at 191M and 374M.} ClimbMix loss
(nats, lower is better) after 20~TPP. Each seed sets
initialization and data order; difference rows compare models within a seed.
Seed 42 is the run reported in the main text. Bold marks the best model in each column
within each scale; shading marks Loopback.}
\label{tab:seeds}
\centering\small
\setlength{\tabcolsep}{4pt}
\renewcommand{\arraystretch}{1.08}
\begin{tabularx}{0.75\linewidth}{@{}l*{2}{>{\raggedleft\arraybackslash}X}@{}}
\toprule
& \multicolumn{2}{c}{Validation loss $\downarrow$} \\
\cmidrule(lr){2-3}
Model & seed 42 & seed 1337 \\
\midrule
\multicolumn{3}{@{}l}{\color{paperNavy}\textbf{191M, 20 layers}} \\
\midrule
Standard & 2.6234 & \textbf{2.6144} \\
Plain Loop & 2.6802 & 2.6815 \\
mHC & 2.6250 & 2.6292 \\
HC & 2.6449 & 2.6427 \\
iHC & 2.6564 & 2.6497 \\
Hyperloop & 2.6489 & 2.6472 \\
MHAR & 2.6564 & 2.6573 \\
\rowcolor{paperTeal!7}
\textbf{Loopback} & \textbf{2.6187} & 2.6155 \\
\midrule
Loopback $-$ Standard & $-0.0047$ & $+0.0011$ \\
Loopback $-$ mHC & $-0.0063$ & $-0.0137$ \\
\midrule
\multicolumn{3}{@{}l}{\color{paperNavy}\textbf{374M, 24 layers}} \\
\midrule
Standard & 2.4198 & 2.4257 \\
mHC & 2.4062 & 2.4079 \\
Hyperloop & 2.4179 & 2.4175 \\
\rowcolor{paperTeal!7}
\textbf{Loopback} & \textbf{2.3971} & \textbf{2.3996} \\
\midrule
Loopback $-$ Standard & $-0.0227$ & $-0.0261$ \\
Loopback $-$ mHC & $-0.0091$ & $-0.0083$ \\
\bottomrule
\end{tabularx}
\end{table}

\noindent\begin{minipage}{\linewidth}
\subsection{Memory-reader results}
\label{app:reader-results}

% Supplementary memory readers at two scales. Included from sections/appendix.tex; uses booktabs,
% cleveref, amsmath and colortbl, all loaded by main.tex.
%
% Protocol, identical for every row of a scale: warmup-stable-decay schedule (linear warmup 128 steps, peak learning
% rate held, last 10% of the budget decayed to zero along 1-sqrt), AdamW with betas 0.9/0.95, eps 1e-15, gradient
% clipping 1.0, no weight decay, sequence length 4096, ClimbMix, seed 42, 20 TPP against the dense control's total
% parameters, validation on the training packing (corrected protocol). 191M (d768 x 20, [2->4x4->2]): looped lr
% 3e-3, dense 1e-3, N4-3, bd_/la_ <kind>_p2k4l4c2 (*_climbval_bos.json). 374M (d1024 x 24, [2->10x2->2]): lr 2e-3
% for every arm, N4-3 [lin58], t24_bd_<kind>_p2k10l2c2 (updates) and t24_la_<kind>_p2k10l2c2 (states) rung 20 (7120
% steps, *_climbval.json, packing=train). The 374M States column was added 2026-10-03 (all eleven state arms finished
% 2026-09-26 under the identical recipe; the arXiv version carries it).
\captionof{table}{\textbf{Memory readers in looped Transformers.} ClimbMix validation loss
($\downarrow$), with matched machine and budget within each scale
(\cref{app:190m-settings,app:learning-rates}). Bold marks the best reader in each column
and the best reference model at each scale. Updates and states use one and
eight heads, respectively, with different output blends (\cref{app:readers}).}
\label{tab:readers-compressed}
\centering
\begin{minipage}{0.84\linewidth}
\centering\small
\begingroup
\setlength{\tabcolsep}{10pt}
\renewcommand{\arraystretch}{1.12}
\begin{tabular*}{\linewidth}{@{\extracolsep{\fill}}lrr@{\hskip 24pt}rr@{}}
\toprule
 & \multicolumn{2}{c}{\textbf{191M}} & \multicolumn{2}{c@{}}{\textbf{374M, 24 layers}} \\
\cmidrule(lr{12pt}){2-3}\cmidrule(l){4-5}
\textbf{Reader} & Updates & States & Updates & States \\
\midrule
Gated DeltaNet & \textbf{2.6489} & 2.6685 & 2.4308 & 2.4333 \\
HGRN-2 & 2.6522 & 2.6619 & \textbf{2.4256} & 2.4333 \\
RWKV-7 & 2.6532 & 2.6791 & 2.4371 & 2.4398 \\
Mesa & 2.6539 & \textbf{2.6603} & 2.4277 & 2.4342 \\
DeltaNet & 2.6542 & 2.6664 & 2.4310 & 2.4329 \\
KDA & 2.6543 & 2.6612 & 2.4290 & \textbf{2.4295} \\
Mamba-2 & 2.6544 & 2.6759 & 2.4308 & 2.4319 \\
LaCT-inspired & 2.6558 & 2.6774 & 2.4308 & 2.4336 \\
GLA & 2.6640 & 2.6773 & 2.4352 & 2.4419 \\
RetNet & 2.6650 & 2.6827 & 2.4338 & 2.4396 \\
Linear attention & 2.6697 & 2.6846 & 2.4346 & 2.4417 \\
\midrule
\multicolumn{5}{@{}l}{\textit{Reference models}} \\[1pt]
Plain Loop & \multicolumn{2}{c}{2.6802} & \multicolumn{2}{c@{}}{2.4415} \\
Standard & \multicolumn{2}{c}{2.6234} & \multicolumn{2}{c@{}}{2.4198} \\
\textbf{Loopback} & \multicolumn{2}{c}{\textbf{2.6187}} & \multicolumn{2}{c@{}}{\textbf{2.3971}} \\
\bottomrule
\end{tabular*}
\endgroup
\end{minipage}
\end{minipage}\par

\FloatBarrier
\Needspace{24\baselineskip}
\section{Depth sweep at fixed parameters}
\label{app:depth}

\Cref{tab:depth-shapes} gives the configurations and results for all nine
rules, using the recipes in \cref{app:larger-settings}. Width varies with
depth to keep Standard's total parameters within 5\% of its 24-layer reference; training
tokens remain fixed. Every run uses one node with eight 80\,GB GPUs:
H100s at 374M and A100s at 778M, including the 48-layer models in
\cref{tab:downstream-main}. Learning rates are $2\times10^{-3}$ for all
374M arms, and $10^{-3}$ for Standard versus $1.5\times10^{-3}$ for looped
arms at 778M. We also check Standard's rate: at 374M, $10^{-3}$ and
$3\times10^{-3}$ underperform $2\times10^{-3}$ after 5~TPP at 48 and 64
layers ($3\times10^{-3}$ diverges at 48). At 778M, using $1.5\times10^{-3}$
changes its 20-TPP loss by $-0.0045$, $-0.0020$, $-0.0010$, and $+0.0076$
at 16, 24, 32, and 48 layers.

% Numeric data originate in figure/depth_sweep/depth_sweep_data.json (outside this repository).
% The combined layout is maintained here. The 374M 24-layer row comes from app:1024-settings.

\begin{table}[H]
\caption{\textbf{Depth-sweep configurations and validation loss at 20~TPP.}
Training tokens are fixed; Standard's total parameters stay within 5\% of its 24-layer reference.
(a) Configurations: heads are 64-dimensional and looped models make two visits.
Transformer counts exclude embeddings and the vocabulary head; shared weights are counted once.
Other looped arms are within 0.7\% of Loopback's Transformer count.
(b) ClimbMix validation loss: bold marks the lowest loss per row; shading marks Loopback.}
\label{tab:depth-shapes}
\label{tab:depth-loss}
\centering\footnotesize
\setlength{\tabcolsep}{4pt}
\textbf{(a) Configurations}\par\smallskip
\begin{tabular*}{\linewidth}{@{\extracolsep{\fill}}lrrcrrrr@{}}
\toprule
& & & & & \multicolumn{2}{c}{Transformer (M)} & \\
\cmidrule(lr){6-7}
Layers & Width & Heads & Looped topology & Standard total (M) & Standard & Loopback & Share \\
\midrule
\multicolumn{8}{@{}l}{\color{paperNavy}\textbf{374M, 7.47B tokens at 20~TPP (7,120 steps of 1,048,576 tokens)}} \\
16 & 1216 & 19 & $\looptopology{1}{7}{2}{1}$ & 366.75 & 288.92 & 162.54 & 56.3\% \\
24 & 1024 & 16 & $\looptopology{2}{10}{2}{2}$ & 373.87 & 308.33 & 179.88 & 58.3\% \\
32 & 896 & 14 & $\looptopology{2}{14}{2}{2}$ & 380.36 & 323.02 & 181.72 & 56.3\% \\
48 & 768 & 12 & $\looptopology{3}{21}{2}{3}$ & 388.97 & 339.81 & 191.18 & 56.3\% \\
64 & 640 & 10 & $\looptopology{4}{28}{2}{4}$ & 366.10 & 325.14 & 182.93 & 56.3\% \\
\addlinespace[2pt]
\multicolumn{8}{@{}l}{\color{paperNavy}\textbf{778M, 15.60B tokens at 20~TPP (7,440 steps of 2,097,152 tokens)}} \\
16 & 1856 & 29 & $\looptopology{1}{7}{2}{1}$ & 795.44 & 676.65 & 380.64 & 56.3\% \\
24 & 1536 & 24 & $\looptopology{2}{10}{2}{2}$ & 777.86 & 679.55 & 396.44 & 58.3\% \\
32 & 1344 & 21 & $\looptopology{2}{14}{2}{2}$ & 779.74 & 693.72 & 390.25 & 56.3\% \\
48 & 1088 & 17 & $\looptopology{3}{21}{2}{3}$ & 778.31 & 708.68 & 398.68 & 56.3\% \\
64 & 960 & 15 & $\looptopology{4}{28}{2}{4}$ & 769.35 & 707.91 & 398.25 & 56.3\% \\
\bottomrule
\end{tabular*}
\par\medskip
\textbf{(b) Validation loss}\par\smallskip
\setlength{\tabcolsep}{2.3pt}
\renewcommand{\arraystretch}{1.12}
\begin{tabular*}{\linewidth}{@{\extracolsep{\fill}}lrrrrrrrr>{\columncolor{paperTeal!7}}r@{}}
\toprule
& & & \multicolumn{4}{c}{Hyper-connection family} & \multicolumn{2}{c}{Attention family} & \\
\cmidrule(lr){4-7}\cmidrule(lr){8-9}
Layers & Standard & Plain Loop & mHC & HC & iHC & Hyperloop & MHAR & AttnRes & \textbf{Loopback} \\
\midrule
\multicolumn{10}{@{}l}{\color{paperNavy}\textbf{374M: ClimbMix loss $\downarrow$, 20~TPP}} \\
16 & 2.4218 & 2.4536 & 2.4234 & 2.4406 & 2.4355 & 2.4327 & 2.4453 & 2.4312 & \textbf{2.4191} \\
24 & 2.4198 & 2.4415 & 2.4062 & 2.4249 & 2.4217 & 2.4179 & 2.4223 & 2.4065 & \textbf{2.3971} \\
32 & 2.4395 & 2.4371 & 2.4118 & 2.4273 & 2.4199 & 2.4147 & 2.4310 & 2.4166 & \textbf{2.4008} \\
48 & 2.4544 & 2.4425 & 2.4089 & 2.4327 & 2.4297 & 2.4154 & 2.4360 & 2.4168 & \textbf{2.4005} \\
64 & 2.4958 & 2.4695 & 2.4328 & 2.4537 & 2.4607 & 2.4346 & 2.4515 & 2.4415 & \textbf{2.4224} \\
\addlinespace[3pt]
\multicolumn{10}{@{}l}{\color{paperNavy}\textbf{778M: ClimbMix loss $\downarrow$, 20~TPP}} \\
16 & 2.2604 & 2.2857 & 2.2626 & 2.2741 & 2.2613 & 2.2696 & 2.2739 & 2.2735 & \textbf{2.2583} \\
24 & 2.2441 & 2.2713 & 2.2432 & 2.2579 & 2.2461 & 2.2551 & 2.2595 & 2.2495 & \textbf{2.2425} \\
32 & 2.2471 & 2.2737 & 2.2471 & 2.2627 & 2.2498 & 2.2536 & 2.2612 & 2.2453 & \textbf{2.2415} \\
48 & 2.2644 & 2.2737 & 2.2454 & 2.2642 & 2.2523 & 2.2539 & 2.2646 & 2.2529 & \textbf{2.2404} \\
64 & 2.2788 & 2.2816 & \textbf{2.2495} & 2.2684 & 2.2580 & 2.2614 & 2.2738 & 2.2507 & 2.2512 \\
\bottomrule
\end{tabular*}
\end{table}

\clearpage
\section{Mathematical and code text}
\label{app:mathreason}

We use these evaluations to test whether parameter-efficient reuse also
preserves mathematical and code solution modeling. The 1.4B and 3B results
are in \cref{tab:scale-large-math}; \cref{tab:mathreason} compares all nine
designs at 191M and 374M. These use the checkpoints and matched training
budgets of \cref{tab:downstream-main}, including the 32-layer 374M setting.

\paragraph{Reference-solution loss.}
We evaluate GSM8K test (1,319 problems), MATH-500 (500), HumanEval (164), and
sanitized MBPP test (257). Each prompt contains a beginning-of-sequence
token and the problem; MBPP additionally includes the first test.
GSM8K calculator annotations are removed and the final line is standardized
to \emph{The answer is $N$.} We score the reference solution with teacher forcing
in a 4,096-token window using BF16 and the reference Loopback read, averaging
NLL over solution tokens. The four-set mean weights datasets equally.
Paired bootstrap resampling uses 2,000 draws over problems.
This metric evaluates solution likelihood, not generated-answer or
code-execution correctness.

\paragraph{Five-shot numeric-answer accuracy.}
The six sets are SingleOp (562), SingleEq (508), AddSub (395), MultiArith
(600), the numeric-answer subset of ASDiv (2,084), and SVAMP (1,000), from
the ToRA evaluation files. We use the first five standard-prompting exemplars
of chain-of-thought prompting, formatted as
\emph{Q: \ldots\ A: The answer is $N$.}
A response is correct when each reference-number token is the greedy choice
and the greedy token following the number is not a digit. The reported mean
weights all six sets equally; the 1.4B and 3B means appear in
\cref{tab:scale-large-math}. At 3B, Loopback is higher on each set:
11.9 versus 4.3 on SingleOp, 8.5 versus 3.3 on SingleEq, 5.6 versus 2.0 on
AddSub, 1.3 versus 0.8 on MultiArith, 7.0 versus 3.4 on ASDiv, and 4.1 versus
1.9 on SVAMP. The low absolute accuracies limit claims about reliable
mathematical problem solving.

% The math-and-code table (tab:mathreason): reference-solution NLL of all nine rules at 191M and 374M,
% generated by figure/mathreason/make_mathreason_tables.py (outside this repository) from the result JSONs in
% figure/mathreason/data/; regenerate instead of editing by hand.
\begin{table}[H]
\setlength{\abovecaptionskip}{4pt}
\caption{\textbf{Reference solutions at matched tokens.} Teacher-forced loss (nats per token, lower is better) of the
reference solutions of GSM8K, MATH-500, HumanEval and MBPP given the problem, for all nine designs
at 191M and 374M. Every prompt starts with the beginning-of-sequence token.
Paired bootstrap standard errors of the differences between rules are at most 0.003 on the mean. Bold marks the lowest
loss per row.}
\label{tab:mathreason}
\centering\footnotesize
\setlength{\tabcolsep}{2.3pt}
\renewcommand{\arraystretch}{0.95}
\begin{tabular*}{\linewidth}{@{\extracolsep{\fill}}lrrrrrrrr>{\columncolor{paperTeal!7}}r@{}}
\toprule
& & & \multicolumn{4}{c}{Hyper-connection family} & \multicolumn{2}{c}{Attention family} & \\
\cmidrule(lr){4-7}\cmidrule(lr){8-9}
Reference solutions $\downarrow$ & Standard & Plain Loop & mHC & HC & iHC & Hyperloop & MHAR & AttnRes & \textbf{Loopback} \\
\midrule
\multicolumn{10}{@{}l}{\color{paperNavy}\textbf{191M, 20 layers, $\looptopology{2}{4}{4}{2}$ (3.82B tokens)}} \\
\midrule
\quad GSM8K & 1.822 & 1.865 & 1.795 & 1.801 & 1.826 & 1.833 & 1.826 & 1.797 & \textbf{1.768} \\
\quad MATH-500 & 1.967 & 2.028 & \textbf{1.961} & 1.999 & 2.023 & 2.024 & 1.994 & 1.962 & 1.972 \\
\quad HumanEval & \textbf{1.827} & 1.935 & 1.861 & 1.900 & 1.924 & 1.905 & 1.897 & 1.895 & 1.842 \\
\quad MBPP & 2.021 & 2.118 & 2.028 & 2.042 & 2.081 & 2.074 & 2.076 & 2.064 & \textbf{2.018} \\
\textbf{Mean} & 1.909 & 1.987 & 1.911 & 1.935 & 1.964 & 1.959 & 1.948 & 1.929 & \textbf{1.900} \\
\midrule
\multicolumn{10}{@{}l}{\color{paperNavy}\textbf{374M, 32 layers, $\looptopology{2}{14}{2}{2}$ (7.47B tokens)}} \\
\midrule
\quad GSM8K & 1.600 & 1.612 & 1.564 & 1.596 & 1.554 & 1.552 & 1.592 & 1.563 & \textbf{1.529} \\
\quad MATH-500 & 1.755 & 1.755 & 1.740 & 1.747 & 1.728 & 1.737 & 1.746 & 1.734 & \textbf{1.707} \\
\quad HumanEval & 1.518 & 1.520 & 1.487 & 1.514 & 1.495 & 1.493 & 1.503 & 1.481 & \textbf{1.447} \\
\quad MBPP & 1.690 & 1.707 & 1.664 & 1.684 & 1.661 & 1.680 & 1.691 & 1.675 & \textbf{1.643} \\
\textbf{Mean} & 1.641 & 1.648 & 1.614 & 1.635 & 1.610 & 1.615 & 1.633 & 1.613 & \textbf{1.582} \\
\bottomrule
\end{tabular*}
\end{table}

\clearpage
\FloatBarrier
\section{3B task results with one extra visit}
\label{app:extra-visits}

We evaluate the same 3B Loopback checkpoint at its trained two visits and
at three visits without retraining. Additional destinations reuse
the last trained visit's query for the same block position; the shared blocks
and coda are unchanged. We report all thirteen zero-shot tasks using the main
evaluation protocol: HellaSwag, OpenBookQA, ARC-Easy, and ARC-Challenge use
length-normalized accuracy; the other tasks use accuracy. The mean weights
the thirteen tasks equally, with MMLU's 57 subjects counted as one task.

% Generated by figures/make_extra_visit_comparison.py from extra_visit_3b_tasks.json.
\begin{table}[H]
\caption{\textbf{3B Loopback with one extra visit.} Zero-shot scores (\%)
at two visits (trained) and three visits, without retraining. $\Delta$ is
the change in percentage points computed from the displayed scores;
bold marks the higher score in each row.}
\label{tab:extra-visits-3b-tasks}
\centering\small
\setlength{\tabcolsep}{10pt}
\renewcommand{\arraystretch}{1.06}
\begin{tabular}{lrrr}
\toprule
Task & 2 visits (trained) & 3 visits & $\Delta$ (pp) \\
\midrule
PIQA & \textbf{75.90} & 75.30 & -0.60 \\
HellaSwag & 63.72 & \textbf{65.11} & +1.39 \\
WinoGrande & 58.48 & \textbf{58.96} & +0.48 \\
OpenBookQA & 41.80 & \textbf{42.80} & +1.00 \\
CommonsenseQA & 21.87 & \textbf{22.36} & +0.49 \\
BoolQ & 64.95 & \textbf{65.60} & +0.65 \\
RACE & \textbf{36.17} & 35.60 & -0.57 \\
ARC-Easy & \textbf{71.13} & 71.04 & -0.09 \\
ARC-Challenge & \textbf{40.96} & 40.02 & -0.94 \\
MMLU (57 subjects) & 28.86 & \textbf{29.69} & +0.83 \\
ANLI-R1 & 32.90 & \textbf{33.70} & +0.80 \\
RTE & 56.68 & \textbf{63.90} & +7.22 \\
LAMBADA & \textbf{49.45} & 48.09 & -1.36 \\
\midrule
Thirteen-task mean & 49.45 & \textbf{50.17} & +0.72 \\
\bottomrule
\end{tabular}
\end{table}

The mean rises by 0.72 points, with gains on eight of thirteen tasks.
RTE (277 items) contributes 0.56 points to the mean increase. These are
single-checkpoint measurements; the improvement is not uniform across tasks.

\clearpage
\section{Compute and efficiency measurements}
\label{app:compute}

\paragraph{FLOP accounting.}
\Cref{tab:compute} covers three scales. PyTorch's operator counter measures
forward FLOPs on one 4,096-token bf16 sequence, including the vocabulary head.
Scaled-dot-product attention substitutes for the uncounted flash kernel;
uncounted elementwise work is added analytically. For a read at destination
$s$ over $s+1$ stored updates, the score and weighted-sum matmuls contribute
$4(s+1)d$ FLOPs per token, excluding normalization, softmax, and mixture
replacement. Anchored decay costs $3d$ per token at each visit end; AttnRes
reads before token attention and before the MLP in each loop block, and once
at loop exit, over the loop input and up to eight loop states; the stream rules cost
$O(n^2 d)$ per connection for $n=4$ streams.

\paragraph{Training measurements.}
Measurements use one eight-H100 node, FSDP, bf16 compute, and each scale's
micro-batch. Loopback uses fused reads measured in 160-step timing
runs; other entries come from the training logs, including AttnRes in
\cref{fig:efficiency}, which trains with its fused read. For these three scales,
throughput uses median step time, and memory is the trainer's reported peak. Activations dominate
sharded model state here, so fewer stored weights need not lower peak memory.
At 3B, the mean logged training throughputs are 24.7k tokens/s/GPU for
Standard and 23.3k for Loopback, a 6\% gap after rounding. Both runs use the
same eight-H100 node, four 4,096-token sequences per GPU, and 60 gradient
accumulation steps. These means cover the logged intervals of the final
20-TPP training stage.

% Generated by figure/efficiency/make_compute_table.py from the measured training logs (train_steptimes.json,
% tim_fused.json), the FLOPs count (flops_per_token.json) and the single-GPU prefill benchmark (bench3.jsonl).
\begin{table}[H]
\caption{\textbf{Compute and efficiency of the residual rules.}
Parameters are Transformer-only; extra forward FLOPs are relative to Standard.
Percentages are rounded to two decimal places, with the positive 374M MHAR
overhead rounded up to $0.01\%$. Training throughput and peak memory
are per GPU. Prefill uses one H100.}
\label{tab:compute}
\centering\footnotesize
\setlength{\tabcolsep}{3pt}
\renewcommand{\arraystretch}{1.06}
\begin{tabular*}{\linewidth}{@{\extracolsep{\fill}}lrrrrr@{}}
\toprule
Model & \shortstack{Transformer\\params (M)} & Extra FLOPs & Train tok/s/GPU & Peak GiB/GPU & Prefill tok/s  \\
\midrule
\multicolumn{6}{@{}l}{\color{paperNavy}\textbf{191M, $\looptopology{2}{4}{4}{2}$, 3.82B tokens, 16 sequences per GPU}} \\
\midrule
Standard & 141.59 & -- & 222k & 38 & 885k  \\
Plain Loop & 56.64 & +0.00\% & 226k & 38 & 886k  \\
mHC & 56.93 & +0.15\% & 169k & 42 & 594k  \\
HC & 56.65 & +0.07\% & 186k & 42 & 671k  \\
iHC & 57.04 & +0.17\% & 116k & 67 & 370k  \\
Hyperloop & 56.79 & +0.08\% & 186k & 42 & 743k  \\
MHAR & 56.64 & +0.01\% & 204k & 41 & 810k  \\
\textbf{Loopback} & 56.65 & +0.08\% & 167k & 42 & 601k  \\
\midrule
\multicolumn{6}{@{}l}{\color{paperNavy}\textbf{374M, $\looptopology{2}{10}{2}{2}$, 7.47B tokens, 8 sequences per GPU}} \\
\midrule
Standard & 308.33 & -- & 128k & 26 & 492k  \\
Plain Loop & 179.86 & +0.00\% & 130k & 26 & 491k  \\
mHC & 180.06 & +0.05\% & 115k & 29 & 422k  \\
HC & 179.87 & +0.02\% & 120k & 30 & 439k  \\
iHC & 180.53 & +0.15\% & 68k & 52 & 220k  \\
Hyperloop & 179.96 & +0.03\% & 120k & 29 & 457k  \\
MHAR & 179.86 & +0.01\% & 126k & 27 & 480k  \\
\textbf{Loopback} & 179.88 & +0.08\% & 98k & 32 & 352k  \\
\midrule
\multicolumn{6}{@{}l}{\color{paperNavy}\textbf{778M, $\looptopology{2}{8}{2}{6}$, 15.60B tokens, 8 sequences per GPU}} \\
\midrule
Standard & 679.55 & -- & 75k & 38 & 269k  \\
Plain Loop & 453.04 & +0.00\% & 76k & 37 & 269k  \\
mHC & 453.33 & +0.04\% & 68k & 42 & 240k  \\
\textbf{Loopback} & 453.06 & +0.04\% & 63k & 44 & 217k  \\
\bottomrule
\end{tabular*}
\end{table}

\Needspace{10\baselineskip}
\paragraph{Inference measurements.}
One H100 runs each model eagerly with bf16 weights and no gradients; Loopback
uses fused reads. Prefill is one forward pass over a batch of
eight 4,096-token sequences that keeps only the last position's logits and
materializes no KV cache, so its measured peak covers weights and transient
activations. For Loopback these include the depth history, which exists only
within a forward pass and is never cached across decoding steps. Loopback's
prefill throughput relative to the standard Transformer is 0.68, 0.72, and 0.81 at 191M, 374M,
and 778M. \Cref{tab:inference-memory} instead counts \emph{persistent} memory
analytically: all stored weights plus the KV cache, whose capacity is equal
at matched executed depth. The two views differ by
construction: at 374M the measured prefill peak is 3.0\,GiB for Loopback
versus 2.1\,GiB for the standard Transformer, whereas the persistent total for the same eight
sequences is 3.46 versus 3.70\,GiB.

% Generated by figure/efficiency/make_inference_memory_table.py (analytic; parameter counts from flops_per_token.json and
% figures/scaling_results.csv, architecture fields from the LoopResidual model configs). 1.4B Loopback = [3->8x2->6]
% (lt2048_ours_p3k8l2c6, 2026-10-03); the 3B rows (2026-10-26) use the same formulas by hand: weights = total params x 2 bytes,
% KV = 2 x 24 executed blocks x 3,072 x 2 bytes x 4,096 tokens = 1,152 MiB (configs_d3072x24, untied head); the caption here is the manuscript's, not the generator's.
\begin{table}[H]
\caption{\textbf{Persistent inference memory of Standard and Loopback.}
Analytical bf16 storage for all weights, including embeddings and the vocabulary
head, plus the KV cache at 4,096 tokens. Totals exclude temporary activations
and workspaces; they are not measured peaks. Shading marks Loopback.}
\label{tab:inference-memory}
\centering\footnotesize
\setlength{\tabcolsep}{3pt}
\renewcommand{\arraystretch}{1.06}
\begin{tabular*}{\linewidth}{@{\extracolsep{\fill}}llrrrrr@{}}
\toprule
& & & & KV cache per & \multicolumn{2}{c@{}}{Weights + KV (GiB)} \\
\cmidrule(l){6-7}
Scale & Model & Total params (M) & Weights (GiB) & sequence (MiB) & 1 sequence & 8 sequences \\
\midrule
191M, 20 layers & Standard & 190.74 & 0.36 & 240 & 0.59 & 2.23 \\
\rowcolor{paperTeal!7}
\cellcolor{white} & \textbf{Loopback} & 105.80 & 0.20 & 240 & 0.43 & 2.07 \\
\addlinespace[2pt]
374M, 24 layers & Standard & 373.87 & 0.70 & 384 & 1.07 & 3.70 \\
\rowcolor{paperTeal!7}
\cellcolor{white} & \textbf{Loopback} & 245.42 & 0.46 & 384 & 0.83 & 3.46 \\
\addlinespace[2pt]
778M, 24 layers & Standard & 777.86 & 1.45 & 576 & 2.01 & 5.95 \\
\rowcolor{paperTeal!7}
\cellcolor{white} & \textbf{Loopback} & 551.36 & 1.03 & 576 & 1.59 & 5.53 \\
\addlinespace[2pt]
1.4B, 25 layers & Standard & 1,415.68 & 2.64 & 800 & 3.42 & 8.89 \\
\rowcolor{paperTeal!7}
\cellcolor{white} & \textbf{Loopback} & 1,004.64 & 1.87 & 800 & 2.65 & 8.12 \\
\addlinespace[2pt]
3B, 24 layers & Standard & 2,914.67 & 5.43 & 1,152 & 6.55 & 14.43 \\
\rowcolor{paperTeal!7}
\cellcolor{white} & \textbf{Loopback} & 2,235.19 & 4.16 & 1,152 & 5.29 & 13.16 \\
\bottomrule
\end{tabular*}
\end{table}
\FloatBarrier
\Needspace{8\baselineskip}
\paragraph{The fused read.}
\label{app:fused}
Each read uses one Triton kernel, streaming updates once per token tile while
accumulating the scored and prior softmaxes together. Backward saves one score
per token and update rather than normalized keys, then reloads each update
and writes its gradient once. Tests compare the kernel with a direct PyTorch
implementation of \cref{eq:instance}. In FP32, the output, the installed
coefficients, and every gradient (query, updates through the value and key
paths, proposal, input, and previous coefficients) agree to within $10^{-6}$
relative error, measured as the largest absolute difference over the largest
reference magnitude, across widths 768--3,072, 2--20 stored updates, ragged
token counts, and boundary and in-visit reads. With bf16 inputs, both
implementations are compared with an FP64 evaluation of the same rounded
inputs, and the fused error is no larger than the reference's. At zero
queries the output and coefficients match the decay-only reference, which
reduces to Plain Loop only at $\lambda=1$. An end-to-end comparison of the two
implementations on a small model gives bitwise-identical results at initialization and agreement within
$10^{-6}$ after perturbing the queries, forward and backward.
Most remaining step-time overhead is in backward, as each update
receives gradients from later reads. The kernel and tests will be released
with the code.

\FloatBarrier

\end{document}